\documentclass[fleqn,ijoc]{informs4}

\OneAndAHalfSpacedXI 

\usepackage{endnotes}
\let\footnote=\endnote
\let\enotesize=\normalsize
\def\notesname{Endnotes}%
\def\makeenmark{$^{\theenmark}$}
\def\enoteformat{\rightskip0pt\leftskip0pt\parindent=1.75em
	\leavevmode\llap{\theenmark.\enskip}}

\usepackage{graphicx}
\usepackage{eqndefns-left}
\usepackage{multirow}
\usepackage{hhline}
\usepackage{natbib}
\usepackage{graphicx}
\usepackage{bm,array}
\usepackage{comment}
\usepackage{caption}
\usepackage{subcaption}
 \usepackage{makecell}
 \usepackage{multirow}
 \usepackage{float}
 \usepackage{lscape}
 \usepackage{svg}
 \usepackage{framed}
\usepackage{appendix}
\usepackage{color}
\definecolor{strcolor}{rgb}{0.6, 0.2, 0.6}
\definecolor{commentcolor}{rgb}{0.3125, 0.5, 0.3125}
\definecolor{keycol}{rgb}{0, 0, 1}

\usepackage{bbm}

\usepackage{listings}
\usepackage{url}

\newcommand {\bea}{\begin{eqnarray}}
	\newcommand {\eea}{\end{eqnarray}}

\def\blot{\quad \mbox{$\vcenter{ \vbox{ \hrule height.4pt
				\hbox{\vrule width.4pt height.9ex \kern.9ex \vrule width.4pt}
				\hrule height.4pt}}$}}

\usepackage{natbib}
\bibpunct[, ]{(}{)}{,}{a}{}{,}%
\def\bibfont{\fontsize{8}{9.5}\selectfont}%

\usepackage[ruled,vlined,linesnumbered]{algorithm2e}
\usepackage{algorithmic}
\newcommand{\comments}[1]{{\color{blue}\textit{$\#$ #1}}}

\newcommand{\transpose}{{\mbox{\tiny T}}}

\newcommand{\cU}{{\mathcal{U}}}

\newcommand{\cV}{{\mathcal{V}}}
\newcommand{\cD}{{\mathcal{D}}}
\newcommand{\cK}{{\mathcal{K}}}

\newcommand{\cL}{{\mathcal{L}}}
\newcommand{\cX}{{\mathcal{X}}}
\newcommand{\cF}{{\mathcal{F}}}
\newcommand{\cN}{{\mathcal{N}}}

\newcommand{\cT}{{\mathcal{T}}}

\newcommand{\cC}{{\mathcal{C}}}
\newcommand{\cH}{{\mathcal{H}}}

\newcommand{\bbe}{{\textbf{e}}}

\newcommand{\bB}{\textbf{B}}

\newcommand{\bx}{\textbf{x}}
\newcommand{\by}{\textbf{y}}
\newcommand{\bs}{\textbf{s}}
\newcommand{\bd}{\textbf{d}}

\newcommand{\bc}{\textbf{c}}

\newcommand{\bA}{\textbf{A}}

\newcommand{\bz}{\textbf{z}}
\newcommand{\bb}{\textbf{b}}

\newcommand{\bw}{\textbf{w}}
\newcommand{\br}{\textbf{r}}

\newcommand{\bap}{\pmb{\alpha}}

\newcommand{\btheta}{{\pmb{\theta}}}

\newcommand{\bbR}{\mathbb{R}}

 \newenvironment{proofs}[1][Proof]{\noindent\textbf{#1.} }{\hfill \ \rule{0.5em}{0.5em}}

\TheoremsNumberedThrough     
\ECRepeatTheorems

\EquationsNumberedThrough    

\gdef\AQ#1{}
\gdef\CQ#1{}
\begin{document}
	

	\RUNAUTHOR{Pham et~al.} %

	\RUNTITLE{Competitive Facility Location under  Random Utilities and Routing Constraints}

\TITLE{Competitive Facility Location under  Random Utilities and Routing Constraints}

\ARTICLEAUTHORS{
\AUTHOR{
Hoang Giang Pham \textsuperscript{a,c*}, 
Tien Thanh Dam \textsuperscript{b}, 
Ngan Ha Duong \textsuperscript{b,d}, 
Tien Mai \textsuperscript{c}, 
Minh Hoàng Hà \textsuperscript{b}
}
\AFF{
$^{a}$ORLab, Faculty of Computer Science, Phenikaa University, Ha Dong, Hanoi, VietNam\\
$^{b}$ORLab-SLSCM, Faculty of Data Science and Artificial Intelligence, National Economics University, Hai Ba Trung, Hanoi, VietNam\\
$^{c}$School of Computing and Information Systems, Singapore Management University, 80 Stamford Rd, Singapore 178902\\
$^{d}$Department of Engineering, Computer Science and Mathematics, University of L'Aquila, 67100 L'Aquila AQ, Italy
}
}

\ABSTRACT{In this paper, we study a facility location problem within a competitive market context, where customer demand is predicted by a random utility choice model. Unlike prior research, which primarily focuses on simple constraints such as a cardinality constraint on the number of selected locations, we introduce \textit{routing constraints} that necessitate the selection of locations in a manner that guarantees the existence of a tour visiting all chosen locations while adhering to a specified tour length upper bound. Such routing constraints find crucial applications in various real-world scenarios. The problem at hand features a non-linear objective function, resulting from the utilization of random utilities, together with complex routing constraints, making it computationally challenging. To tackle this problem, we explore three types of valid cuts, namely, outer-approximation and submodular cuts to handle the nonlinear objective function, as well as sub-tour elimination cuts to address the complex routing constraints. These lead to the development of two exact solution methods: a \textit{nested cutting plane} and \textit{nested branch-and-cut} algorithms, where these valid cuts are iteratively added to a master problem through two nested loops. We also prove that our nested cutting plane method always converges to optimality after a finite number of iterations. Furthermore,  we develop a local search-based metaheuristic tailored for solving large-scale instances and show its pros and cons compared to exact methods. Extensive experiments are conducted on problem instances of varying sizes, demonstrating that our approach excels in terms of solution quality and computation time when compared to other baseline approaches.}

\FUNDING{The research of the first author is supported by Phenikaa University under grant number PU2023-1-A-01.}

\SUBJECTCLASS{Outer-approximation; Vehicle routing}

\AREAOFREVIEW{Optimization}

	
\maketitle
	
\section{Introduction}\label{sec:intro}
Facility location has been a longstanding problem in decision-making within modern transportation and logistics systems. In its traditional form, this problem involves choosing a subset of potential locations from a given pool of candidates and deciding on the financial resources allocated to establish new facilities at these selected locations. The goal is to either maximize a profit value (such as expected customer demand or revenue) or minimize a cost metric (such as operational or transportation expenses). The decision of where and how to open these facilities is heavily dependent on customer demand, making it a crucial factor in facility location problems. This study focuses on a specific category of \textit{competitive facility location} problems in which customer demand is characterized and forecasted using a random utility maximization (RUM) model \citep{Trai03,FLO_Benati2002maximum,MaiLodi2020_OA}. In this setting, it is assumed that  customers make their choices among available  facilities by maximizing their utility associated with each facility. These utilities typically hinge on attributes (or features) of the facilities, such as service quality, infrastructure, or transportation costs, or characteristics  of customers (such as age, income, gender). The application of the RUM models framework in this context is well justified by the enduring popularity and proven success of RUM models in modeling and forecasting human choice behavior \citep{Mcfadden1978modeling,Mcfadden2001economicNobel,BenABier99a}. To the best of our knowledge, existing studies on competitive facility location under RUM models typically incorporate simple constraints, such as a capacity constraint that imposes an upper limit on the number of selected locations. In this work, we further advance the relevant literature by considering routing constraints, which necessitate the existence of a tour that visits all selected locations with a length below a specified threshold. Such a constraint holds significant relevance in practical scenarios; for example, a bike-sharing company needs to traverse through all bike station locations daily for bike redistribution, and they need to accomplish this within a reasonable travel time. Another example can be found in the situation where a restaurant chain needs to design an efficient supply chain in which a set of locations is chosen to establish new restaurants. The main objective of the restaurants is to fulfill the customer demand as much as possible. But the restaurant manager also tries to keep the internal freight costs and/or transport time reasonable. As such, a threshold should be imposed on the logistics cost or the traveling time to distribute the goods throughout the chain. The inclusion of these constraints poses several challenges, given the nonlinearity of the objective function and the fact that finding a feasible solution to a routing constraint is already an \textit{NP-hard} problem \citep{Laporte2007}. We address these challenges in the paper.

In the context of competitive facility location under  RUM models, the objective is to maximize the expected captured customer demand (or the expected number of customers choosing the newly opened facilities). 
This problem is often referred to as the maximum capture problem (MCP).
The objective function of the MCP is typically characterized by its fractional structure, resulting from the incorporation of RUM models to predict customer demand. Approaches have been proposed to address this nonlinear structure, including Mixed-Integer Linear Programming (MILP) or quadratic second-order cone (Conic) reformulation \citep{FLO_Haase2014comparison,Sen2017}, cutting plans, and branch-and-cut methods using outer-approximation and/or submodular cuts \citep{Ljubic2018outer,MaiLodi2020_OA}. As mentioned earlier, existing approaches only target problems with simple constraints (e.g., cardinality constraints on the number of selected locations) and would be inefficient in handling complex routing constraints. On the other hand, routing constraints have been considered in classical facility location problems where the objective function is linear, making existing approaches inapplicable \citep{GUNAWAN2016}. In this work, to efficiently solve the problem of interest, we will exploit various valid cuts that can be used in a cutting plane and Branch-and-Cut (B\&C) methods to exactly solve the problem. Heuristic methods are also explored to deal with large-scale instances. We present our detailed contributions in the following.

\paragraph{Our contributions:} We make the following  contributions:
\begin{itemize}
   \item[(i)] To efficiently solve the facility location problem at hand, we explore three types of valid cuts that can be employed in cutting plane or Branch-and-Cut (B\&C) methods, namely, outer-approximation, submodular, and sub-tour elimination cuts. While submodular and outer-approximation cuts can be utilized to approximate the nonlinear objective function, sub-tour elimination cuts prove valuable for eliminating infeasible routing solutions that contain sub-tours derived from solving a relaxation problem. To efficiently utilize both outer-approximation and submodular cuts, we introduce the concept of valid cuts that unites both types under a common umbrella. We then demonstrate that a cutting plane method with valid cuts consistently yields an optimal solution after a finite number of iterations. It is noteworthy our finding is more general than existing results,  as that prior work has only demonstrated convergence for cutting plane with outer-approximation cuts.
    \item[(ii)] We further develop a new algorithm, called the nested cutting plane (NCP) algorithm, that allows for incorporating both valid cuts and sub-tour elimination cuts in a nested manner. Our algorithm works by iteratively running two nested loops, where the outer loop involves adding outer-approximation and sub-modular cuts to a master problem, and the inner loop involves iteratively solving and adding sub-tour elimination cuts to the master problem of the outer loop until obtaining a feasible routing solution. We then demonstrate that our nested procedure will always terminate at an optimal solution after a finite number of iterations.
    \item[(iii)] We extend the idea to the context of B\&C and develop a Nested B\&C algorithm that works in the same spirit. That is, at each node of the B\&C, if sub-tours exist, we add sub-tour elimination cuts to remove this infeasible solution, until a feasible routing solution is found. Once this is achieved, we add outer-approximation and sub-modular cuts to further approximate the objective function.
    \item[(iv)] In order to obtain effective solutions for large-scale instances, we further propose a local search-based metaheuristic. It is important to highlight that our algorithm holds distinct advantages over certain prior local search algorithms developed for the competitive facility location problem, particularly in efficiently addressing complex routing constraints.
    \item[(v)] We conduct extensive experiments using instances of various sizes, including those obtained from a real Park-and-Ride dataset. We compare our cutting plane and B\&C methods with several baselines, including the MILP and Conic reformulations, as well as other cutting plane and B\&C versions. The comparison results show that our approach significantly outperforms other baselines in terms of both solution quality and runtime.
\end{itemize}

\paragraph{Paper Outline:} The paper is organized as follows. Section \ref{sec:review} presents a literature review. Section \ref{sec:prob.formulation} presents the problem formulation, and Section \ref{sec:MIP and Conic} discusses MILP and Conic reformulations. In Section \ref{sec: CP and BC}, we present our cutting plane and B\&C approaches, and Section \ref{sec:Heuristic} describes our metaheuristic. Section \ref{sec:Experiment} provides numerical experiments, and finally, Section \ref{sec:concl} concludes.

\section{Literature Review}\label{sec:review}
The variant of the MCP proposed in this paper is an extension of the facility location under random utility choice models. It is also closely related to the Orienteering Problem (OP) \citep{Golden1987}, a well-known variant of the classical Traveling Salesman Problem (TSP), which revolves around optimizing the choice of locations while considering routing constraints. In the context of OP, the usual setting involves each location being assigned a constant score, and the objective is to maximize the total score of the selected locations. In the following, we review the main works studying these two related problems.
\paragraph{Facility location under random utility choice models:} In the context of the MCP, a majority of existing studies employ the MNL model to model customers' demand due to its simple structure. For instance, \cite{FLO_Benati2002maximum} seem to be the first to propose the MCP under the MNL model. Their solution method involves a MILP method based on a branch-and-bound procedure to solve small instances and a simple variable neighborhood search to deal with larger instances. Subsequently, alternative MILP models are introduced by \cite{Zhang2012} and \cite{Haase2009}. \cite{FLO_Haase2014comparison} benchmark these MILP models and conclude that the formulation of \cite{Haase2009} exhibits the best performance. \cite{Freire2015} strengthen the MILP formulation of \cite{Haase2009} using a branch-and-bound algorithm with tight inequalities. \cite{Ljubic2018outer} propose a B\&C method that combines outer-approximation and submodular cuts, while \cite{MaiLodi2020_OA} develop a multicut outer-approximation algorithm for efficiently solving large instances. We note that the MNL model is restricted due to its independence from the Independence from Irrelevant Alternatives (IIA) property, which implies that the ratio of the probabilities of choosing two facilities is independent of any other alternative. This property is shown to not hold in many practical contexts, and efforts have been made to overcome this. Some examples are the mixed logit model (MMNL) \citep{McFaTrai00}, models belonging to the family of Generative Extreme Value (GEV) model such as the nested logit \citep{BenA73,BenALerm85} and the cross-nested logit \citep{VovsBekh98}.  All of the aforementioned studies utilize either the MNL or Mixed Logit (MMNL) models, leveraging the linear fractional structures of the objective functions to devise solution algorithms. \cite{dam2022submodularity,dam2023robust} make the first attempt to incorporate general GEV models into the MCP by proposing a heuristic method that dominates existing exact methods. Recently, \cite{Lamontagne2023} propose a new formulation using the MCP with random utilities to represent the decision of the users to purchase electric vehicles (EVs). The authors determine the optimal placements of EV charging stations in a long-term plan to maximize the total number of EVs by using the simulation-based approach, the rolling horizon heuristic, the greedy methods, and the GRASP algorithms. As far as our understanding goes, none of the existing studies takes into account the routing constraints for the chosen locations. Our research appears to be the first to address routing constraints in the context of the MCP.

\paragraph{Orienteering problem:} Several surveys on the OP, such as \cite{Feillet2005}, \cite{Laporte2007}, \cite{VANSTEENWEGEN2011}, and \cite{GUNAWAN2016}, present a comprehensive summary of the OP and its variants. These surveys mention the Team OP, the (Team) OP with time windows, the Time-dependent OP, the TSP (or Vehicle Routing Problem - VRP) with profits, Prize-collecting TSP (VRP), the Profitable tour, the OP with hotel selection, the Stochastic (Team) OP, the Generalized OP, the Arc OP, the Multi-agent OP, the Clustered OP, the Correlated OP, and others. Besides those problems, \cite{GUNAWAN2016} also review several practical applications of the OP, such as the mobile crowd-sourcing problem, the tourist trip design problem, the theme park navigation problem, the sales representative planning problem, the smuggler search problem, the wildfire routing problem, and the integration of vehicle routing, inventory management and customer selection problems.

Currently, there is still ongoing interest in various variants and applications of the OP within the research community. Following the most recent survey by \cite{GUNAWAN2016}, numerous new OP variants and applications have been proposed. For instance, \cite{ANGELELLI2017} introduce the Probabilistic OP, where each arc is associated with a cost, and nodes have a prize and a specified probability of being visited. Additionally, \cite{ARCHETTI2018} pioneer the Set OP, a generalization of the OP where nodes are organized into clusters, and each cluster is associated with a profit. Notably, in the Set OP, the profit of a cluster can be obtained if at least one node in that cluster is visited, distinguishing it from the Clustered OP. Furthermore, \cite{DOLINSKAYA2018} present a new class of the OP known as the Adaptive OP, which integrates the adaptive path-finding component with the OP into a unified problem. \cite{FREEMAN2018} introduce Attractive OP, considering the competitive factor of each location. Based on that factor, the time-varying attraction of customers to events organized at the chosen location can be determined by a gravity model. Especially, in  Attractive OP, the tour can re-visit a location if it is still one of the locations with the highest attraction after a certain amount of time. The objective of the Attractive OP is to maximize the profit of the tour under the consideration of the total cost for organizing events and traveling among the tour stops. In \cite{Yu2022}, a robust variant of the (team) orienteering problem is proposed. In which, each vehicle serves each customer in an uncertain service time, which is characterized by a capacitated uncertainty set, and archives decreasing profits within a time horizon. Some extensions of the OP are also considered in the fields of robotics, autonomy, and UAV, such as the Surviving Team OP \citep{Jorgensen2017,Jorgensen2018}, the Dubins OP \citep{Penicka2017}, the Close Enough Dubins OP \citep{Faigl2017}, the Dubins Correlated OP \citep{Tsiogkas2018}, the Physical OP \citep{Penicka2019}, and the Kinematic OP \citep{Meyer2022}.

The primary distinction between the MCP and other OP studies lies in the score associated with each location if it is open. The score in the OP is constant. However, in many practical contexts, e.g., the chain of restaurant mentioned above, it is not easy to measure it as a predefined value because of uncertain dependent factors. As such, the score can be estimated via a choice model, which is more pratical. As a consequence, the OP's objective function is linear, whereas the objective function in our new MCP is highly nonlinear, which is much harder to solve. To the best of our knowledge, our work marks the first time addressing the OP with a nonlinear objective function derived from the use of RUM to capture customer demand. In other words, our study represents the first exploration of a combination of competitive facility location problems under RUM and vehicle routing problems.

\section{Maximum Capture Problem with Routing Constraints}\label{sec:prob.formulation}

In the classical facility location problem, decision-makers aim to locate new facilities that maximize the demand captured by the customers. In practice, the customers' demands are difficult to evaluate and not deterministic. In this work, we use discrete choice models to predict customer demand. In the literature on discrete choice models, RUM  \citep{Trai03} is the most widely used approach to model discrete choice behavior. This approach is based on random utility theory, which assumes that the decision makers’ preferences for an alternative are captured by a random utility, and the alternative with the highest utility is chosen by the customer. Under the RUM framework (\cite{McFa78,FosgBier09}), the probability that individual $n$ selects option $i\in S$ can be computed as  $P(u_{ni}\geq u_{nj},\;\forall j\in S)$, i.e., the individual chooses the option of the highest utility, where the random utilities are defined as   $u_{ni}=v_{ni} + \epsilon_{ni}$, in which $v_{ni}$ is a deterministic part and can be modeled based on characteristics of the alternative and/or the decision-maker, and $\epsilon_{ni}$ is the random term which is unknown to the analyst.  Under the MNL model, the choice probability that  the facility at location $i$ is chosen by an individual $n$ is
\[
P_n(i|S) = \frac{e^{v_{ni}}}{\sum_{i\in S}e^{v_{ni}}}
\]


In this paper, we are interested in a competitive facility location problem where a ``\textit{newcomer}'' company wants to open new facilities in a market that already has competitors' facilities. The aim is to maximize the market share by attracting customers to their new facilities. In order to predict how the new facilities affect customer demand, we employ the RUM framework and assume that each customer associates each facility (new facility and one from the competitor) with a random utility and choices are made by maximizing their individual utilities. Accordingly, the company aims to choose a strategic set of locations to establish new facilities to maximize the expected number of customers that would visit the new facilities.
Let us denote $[m] = \{1,2,\ldots,m\}$ as the set of possible locations that can be used to locate new facilities and $\cN$ as the set of geographical zones where customers are located.  The set $\mathcal{N}$ can also be interpreted as a set of customer types, characterized by customers' characteristics such as location, gender, or income. Let $q_n$ be the number of customers at zone $n\in \cN$.  We also denote by $v_{ni}$ the deterministic utility associated with location $i\in [m]$  and zone $n\in \cN$. We can also use $\cC$ to represent the set of the competitor's facilities. 
We now suppose that the firm selects a  subset of locations $S \subset [m]$ to establish new facilities. Under the  MNL  model, the probability that customers in zone $n\in \cN$ select the facility  $i \in S$  can be computed as 
\[
P_n(i|S) = \frac{e^{v_{ni}}}{\sum_{i\in \cC} e^{v_{ni}}+ \sum_{j\in S}e^{v_{nj}}}
\]
 For notational simplicity, let  $U^c_n = \sum_{i\in \cC} e^{v_{ni}}$, noting that these terms are constants in the MCP.  The expected customer demand captured over all the customer zones can be computed as  
 \[
 f(S) = \sum_{n\in \cN} q_n \sum_{i\in S} P_n(i|S) = \sum_{n\in \cN}\frac{\sum_{i\in S}e^{v_{ni}}}{U^c_n + \sum_{i\in S}e^{v_{ni}}}
 \]

Previous studies often consider a simple cardinality constraint $|S| \leq C$ where $C$ is a given constant such that $1\leq C \leq m$ \citep{Freire2015,Ljubic2018outer,MaiLodi2020_OA,dam2022submodularity}. In a practical situation, it is often necessary for selected locations to be efficiently connected to facilitate transportation and logistics among the facilities. For instance, a company may need to regularly visit all of its facilities for purposes like delivering goods or supplies. The previous approaches, which only considered cardinality, do not accommodate this requirement. In the following, we introduce an MCP formulation with routing constraints to overcome this limitation.

We now describe the MCP with a routing constraint (denoted as MCP-R). The MCP-R can be described as a combination of the traditional MCP and OP where the objective is to choose a subset of locations $S \subset [m]$, in a way that maximizes the expected customer demand $f(S)$. This must be achieved while also guaranteeing the existence of a tour that starts and ends at a designated \textit{depot}, and visits all the chosen locations in  $S$ within a specified time limit  $T_{\max}.$ In the most compact form, the MCP-R can be formulated as follows:
\begin{align}
    \max_{S \subset [m]} \qquad & f(S) & \label{prb:MCP-R}\tag{\sf MCP-R}\\
    \text{s.t.} \qquad & |S| \leq C\nonumber\\
    &  \cT(S) \leq T_{\max}\nonumber
\end{align}
where $\cT(S)$ represents the {shortest} possible tour's length that starts and ends at the \textit{depot}, and visits all the nodes in $S$. 
It's important to highlight that without the routing constraint $\mathcal{T}(S) \leq T_{\max}$, any optimal solution to the MCP will essentially meet the maximum possible cardinality, as the objective function $f(S)$ is monotonically increasing. However, with the inclusion of the routing constraint, it is possible for an optimal solution, denoted as $S^*$, not to reach its maximum cardinality size, i.e.,  $|S^*|$ is strictly less than $C$. Moreover, it can be shown that the feasible set of \eqref{prb:MCP-R}  does not form a \textit{matroid}, an important concept in submodular maximization. We state these results in Proposition \ref{prop:matroid} below.

\begin{proposition}\label{prop:matroid}
An optimal solution $S^*$ to \eqref{prb:MCP-R} does not necessarily reach the maximum cardinality size $C$  and the feasible set $\left\{S\big|~|S|\leq C;~ \cT(S) \leq T_{\max}\right\}$ may not form a matroid. 
\end{proposition}

\begin{proofs}
\begin{figure}[htb]
    \centering
    \includegraphics[width=0.4\textwidth]{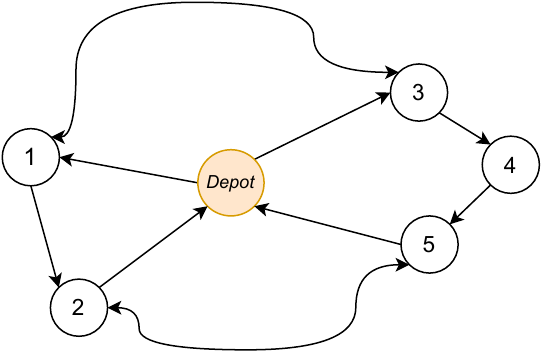}
    \caption{Illustrative example of a situation that the feasible set $\left\{S\big|~|S|\leq C;~ \cT(S) \leq T_{\max}\right\}$ is not a matroid.}
    \label{fig:matroid-example}
\end{figure}
The first claim is easy to verify by using the following counter-example. Let us assume that $C=3$ and among $m$ locations,  there are two locations $\{1,2\}$ near the \textit{depot} such that there is a feasible tour visiting these locations, i.e., $\cT(\{1,2\}) \leq T_{\max}$. We further assume that any other locations are far away from the \textit{depot}, thus including any of them will lead to an infeasible tour.  This implies that if $S^*$ is optimal (and feasible) to \eqref{prb:MCP-R} then $|S^*|\leq 2 <C$ ,  affirming the first claim. 

For the second claim, we first state the definition of \textit{matroid} as follows.   A pair $(\cV,\cX)$, where $\cX\subset 2^m$ is a collection of subsets of $\cV$, is a matroid if it satisfies the following properties:
\begin{itemize}
    \item [(i)] For any $A\subset B \subset \cV$, if $B\in \cX $ then $A \in \cX$
    \item [(ii)] For any $A,B \in \cX$, $|B|>|A|$, then $\exists i \in B\backslash A$ such that $A \cup \{i\} \in \cX$.
\end{itemize}

We now show that the feasible set $\cX = \{S,~ |S|\leq C; \cT(S)\leq T_{\max}\}$ will violate Properties (ii) of the above Matroid definition. To this end, we use a simple example illustrated in Figure \ref{fig:matroid-example}. In this example, assume that there are 5 locations and one \textit{depot}. The locations are divided into two groups: $\{1,2\}$ and $\{3,4,5\}$. We presume that the distance between any pair of nodes from these two groups exceeds $T_{max}$. This implies that a feasible tour cannot encompass locations from both groups. We further assume that $C=3$ and the two tours $\{\text{depot}\rightarrow 1\rightarrow 2 \rightarrow \text{depot}\}$  and $\{\text{depot}\rightarrow 3\rightarrow 4 \rightarrow 5 \rightarrow \text{depot}\}$  are feasible. Consequently,  $A = \{1,2\} \in\cX$ and $B = \{3,4,5\} \in \cX$. Moreover, we see that $|B|>|A|$, but for any location $i \notin A$,  we see that  $A\cup \{i\} \notin \cX$, which is because a feasible tour cannot encompass locations from both $A$ and $B$. As a result, Property (ii) is violated, i.e.,  $(\cV,\cX)$ is not a matroid. \end{proofs}

The concept of matroid plays a pivotal role in submodular maximization -- it has been demonstrated that when dealing with submodular maximization problems subject to matroid constraints, there exists a polynomial-time greedy algorithm that consistently provides at least a 1/2-approximation solution \citep{calinescu2007maximizing}. In the case of the MCP-R, the matroid property does not hold, implying that devising a constant-factor approximation algorithm would be challenging. It is worth noting that the MCP (without routing constraints) is already known to be \texttt{NP-hard}, and verifying whether a solution $S$ satisfies the routing constraint $\cT(S) \leq T_{\max}$ is also an \texttt{NP-hard} problem. Consequently, the MCP-R problem emerges as a particularly challenging problem. 

The routing constraint in \eqref{prb:MCP-R} is impractical. However, by building upon previous research \citep{VANSTEENWEGEN2011,GUNAWAN2016}, it is possible to represent it by a set of linear constraints.  To facilitate this, we first consider the binary representation of the objective function:
\[
    f(\bx) =  \sum_{n\in \cN}\frac{q_n\sum_{i\in [m]}{V_{ni} x_i}}{U^c_n + \sum_{i\in [m]}{V_{ni} x_i}} 
\]
where $V_{ni} = e^{v_{ni}}$ for all $n\in \cN$  and  $i\in [m]$. In other words,  we use $\bx \in \{0,1\}^m$ to represent a subset $S\subset [m]$ as $x_i = 1$ if $i\in S$ and $x_i=0$ otherwise, for all $i\in [m]$.  
To represent the routing constraint as  linear ones, 
we extend the set of locations as $\widetilde{\cV} = \{0,1,...,m,m+1\}$ where Locations 0 and $m+1$ are the $depots$.
We then introduce new decision variables $\by$ such that  $y_{ij} = 1$ if there exists a visit from locations $i$ to location $j$, and 0 otherwise, for all $i,j \in \widetilde{\cV}$. With these new variables, we can reformulate \eqref{prb:MCP-R} as the following mixed-integer nonlinear program:
\begin{align}
    \max_{\bx,\by}& \qquad f(\bx) =  \sum_{n\in \cN}\frac{q_n\sum_{i\in [m]}{V_{ni} x_i}}{U^c_n + \sum_{i\in [m]}{V_{ni} x_i}} & \label{prb:MINLP}\tag{\sf MINLP}\\
    \text{s.t.}  & \qquad \sum_{i\in [m]}x_i \leq C \nonumber\\
    & \qquad  	\sum_{i = 1}^{m+1} y_{0i} = \sum_{i = 0}^{m} y_{i,m+1} = 1  \label{eq:start-end}\\
    &\qquad 	\sum_{j = 0}^{m} y_{ji} = \sum_{k = 1}^{m+1} y_{ik} = x_i \qquad \forall i \in [m]  \label{eq:at-most-once-1} \\
&\qquad  	\sum_{i = 0}^{m} \sum_{j = 1}^{m+1} t_{ij}y_{ij} \leq T_{max} \label{eq:budget}\\
&\qquad p_i - p_j + 1 \leq m(1 - y_{ij}) \qquad \forall i,j \in [m] \label{eq:subtour}\\
 &\qquad 	1 \leq p_i \leq m \qquad \forall i \in [m]\label{eq:bound-p}\\
 &\qquad \bx  \in \{0,1\}^m, \by \in \{0,1\}^{(m+1)\times (m+1)}\nonumber
 \end{align}
where Constraints \eqref{eq:start-end}--\eqref{eq:subtour} are linear representations of the routing constraint $\cT(S) \leq T_{\max}$. More precisely, Constraints (\ref{eq:start-end}) are to guarantee that the route starts from $depot$ $(0)$ and ends on $depot$ $(m+1)$. Constraints (\ref{eq:at-most-once-1}) establish the relationship between variables $\bx$ and $\by$ --  they ensure that a location $i$ is visited iff it is chosen for opening a facility. Constraints (\ref{eq:budget}) ensure that the total traveling time of the selected tour is less than the time budget $T_{max}$, where  $t_{ij}$ be the non-negative travel time between locations $i$ and $j$, $\forall i,j\in [m]$. Finally,  Constraints (\ref{eq:subtour})and (\ref{eq:bound-p}) are the subtour elimination constraints formulated according to the Miller-Tucker-Zemlin (\textbf{MTZ}) formulation of the TSP proposed by \cite{MTZ1960}. 

\section{MILP and Second-order Cone Reformulations}\label{sec:MIP and Conic}

The mixed-integer nonlinear formulation in \eqref{prb:MINLP} is a binary fractional program with a fractional objective function and linear constraints. It is then known that such a fractional program can be reformulated as an MILP or Second-order Cone (or Conic) program, and, as a result, the MCP-R can be readily solved by an off-the-shelf solver such as CPLEX. In the following, we present MILP and Conic reformulations for \eqref{prb:MINLP}. Even though the results are quite straightforward, given existing works, we include them for the sake of self-containment. These reformulations will also serve as our reference benchmarks for comparing against the methods we propose.

\subsection{MILP Reformulation}
We employ the method proposed in \citep{LI1994}, one of the most popular approaches to formulate a binary fractional program into a MILP. In particular, we first introduce new variables $w_n$, $n\in \cN$, as: 
\[ w_{n}= \frac{1}{\sum_{i\in [m]} V_{ni}x_i+U^c_n},~\forall n\in \cN \nonumber\]
\[ r_{n}=\frac{1}{w_{n}},~ \forall n\in \cN \nonumber
\]
Then we can rewrite the objective function as 
\[
f(\bx) = \sum_{n\in \cN}q_n\sum_{i\in [m]} V_{ni} x_i w_n
\]
and the relationship between $w_n$ and $x_i$ can be expressed as 
$w_n(\sum_{i\in [m]} V_{ni}x_i +U^c_n) = 1$. 
The MCP can be formulated as follows:
\begin{align}
    \max_{\bx,\bw, \by} & \qquad \sum_{n\in \cN}q_{n}\sum_{ i\in [m]}V_{ni}x_{i}w_{n}  \label{prb:Li-bilinear}\tag{\sf Li-bilinear}\\
    \text{s.t.} & \qquad \sum_{i\in [m]} x_i \leq C \nonumber \\
    &\qquad \sum_{i\in [m]} V_{ni}x_iw_n + U^c_n w_n = 1 \qquad \forall n\in \cN \label{eq:defw}\\
    &\qquad \text{Routing constraints:  \eqref{eq:start-end}--\eqref{eq:subtour} } \nonumber\\
&\qquad \bx  \in \{0,1\}^m, \by \in \{0,1\}^{(m+1)\times (m+1)}\nonumber    
\end{align}


We observe that the formulation \eqref{prb:Li-bilinear} includes non-convex bilinear terms $x_{i}w_{n}$. \cite{LI1994} suggests such a bilinear term can be linearized by adding variables $s_{ni} = x_{i}w_{n}$ and  replacing the bilinear terms by the following  linear inequalities: (i) $w_{n} - s_{ni} \leq w_{n}^{U} - w_{n}^{U}x_{i}$, (ii) $s_{ni} \leq w_{n}$, (ii) $s_{ni} \leq w_{n}^{U}x_{i}$, where $w_{n}^{U},~ n\in \cN$ are an upper bound of $w_{n}$. Then, \cite{Tawarmalani2002} exploit some properties relating to the convex and concave envelopes of bilinear terms and suggest adding two linear inequalities involving the lower bound of $w_{n}$. In summary,  the following inequalities can be used to represent the terms $x_iw_n$:
\begin{align}
    & s_{ni} \leq w_{n}^{U}x_{i} \qquad \forall i \in [m], n\in \cN \label{eq:linear1}\\
    & w_{n}^{L}x_{i} \leq s_{ni} \qquad \forall i \in [m], n\in \cN \label{eq:linear2}\\
    & s_{ni} \leq w_{n} + w_{n}^{L}(x_{i}-1) \qquad \forall i \in [m], n\in \cN \label{eq:linear3}\\
    & w_{n} + w_{n}^{U}(x_{i}-1) \leq s_{ni} \qquad \forall i \in [m], n\in \cN \label{eq:linear4} 
\end{align}
where $w_{n}^{U}$ and $w_{n}^{L}$ are upper and lower bounds of $w_{n}$.
Finally, we obtain the following MILP reformulation: 
\begin{align}
    \max_{\bx,\bw, \by,\bs} & \qquad \sum_{n\in \cN}q_{n}\sum_{ i\in [m]}V_{ni}s_{ni}  \label{prb:MILP}\tag{\sf MILP}\\
    \text{s.t.} & \qquad \sum_{i\in [m]} x_i \leq C \nonumber \\
    &\qquad \sum_{i\in [m]} V_{ni}s_{ni} + U^c_n w_n = 1 \qquad \forall n\in \cN\\
     &\qquad \text{Constraints:  \eqref{eq:linear1}--\eqref{eq:linear4} } \nonumber\\
    &\qquad \text{Routing constraints:  \eqref{eq:start-end}--\eqref{eq:subtour} } \nonumber\\
&\qquad \bx  \in \{0,1\}^m, \by \in \{0,1\}^{(m+1)\times (m+1)}\nonumber    
\end{align}

\subsection{Conic Reformulation}

Second-order cone (or Conic) optimization refers to the optimization of a linear function over conic quadratic inequalities of the form $||\bA\bz -\bb||_2 \leq \bc \bx + \bd$, where $||\cdot||_2$ is the L2 norm, and $\bA, \bb,\bc,\bd$ are matrices/vectors of appropriate sizes. In the context of 0-1 fractional programming, prior research makes use of the fact that that such a rotated second-order cone/hyperbolic inequality of the form $z_1^2\leq z_2z_3$, for $z_1,z_2,z_3\geq 0$, can be formulated as a conic quadratic inequality  $||(2z_1,z_2-z_3)||\leq z_2+z_3$, and if $z_1$ is a binary variable, then the inequality $z_2z_3\geq z_1$ is equivalent the rotated second-order cone $z_2z_3 \geq z_1^2$.
Note that such a conic quadratic optimization problem can now be solved by some optimization tools such as CPLEX or GUROBI.

In the context of the MCP-R, the methods presented in \cite{Sen2017} and  \cite{Mehmanchi2019} can be directly employed to transform \eqref{prb:MINLP} into a Conic program. Nevertheless, we take advantage of the inherent structure of the MCP to create a simpler Conic formulation. To achieve this, let us first rewrite the objective function in \eqref{prb:MINLP} as a sum of ratios whose numerators are constant.
\begin{align}
    & f(\bx) = \sum_{n\in \cN}\frac{q_n\sum_{i\in [m]}{V_{ni} x_i}}{U^c_n + \sum_{i\in [m]}{V_{ni} x_i}} \nonumber\\
    & = \sum_{n\in \cN}q_n - \sum_{n\in \cN} \frac{q_nU^c_n}{ U^c_n+\sum_{i\in [m]} V_{ni}x_i} \nonumber\\
\end{align}
We also introduce two new sets of variables $w_{n}= \frac{1}{\sum_{i\in [m]} V_{ni}x_i+U^c_n},~\forall n\in \cN$ and $r_{n}=\frac{1}{w_{n}},~ \forall n\in \cN$, then rewrite  \eqref{prb:MINLP} as:
 \begin{align}
    \max_{\bx,\bw, \by,\br} & \qquad \sum_{n\in \cN}q_{n} -  \sum_{n\in \cN}q_n U^c_n w_{n}  \\
    \text{s.t.} & \qquad \sum_{i\in [m]} x_i \leq C \nonumber \\
     &\qquad r_n =  \sum_{i\in [m]}V_{ni}x_i + U^c_n \qquad \forall n\in \cN\nonumber\\
    &\qquad r_n w_n = 1 \qquad \forall n\in \cN \label{eq:cone}\\
    &\qquad \text{Routing constraints:  \eqref{eq:start-end}--\eqref{eq:subtour} } \nonumber\\
&\qquad \bx  \in \{0,1\}^m, \by \in \{0,1\}^{(m+1)\times (m+1)}\nonumber    
\end{align}
 The constraints $w_nr_n = 1$, $n\in \cN$, are not  rotated cones. We however see that these constraints can be safely replaced by $w_nr_n \geq 1$. This is possible because $w_n$  and $r_n$ are non-negative, and the maximization problem will force $w_n$ to be as small as possible (to achieve minimal objective values). So, at optimum, 
 the inequalities  $w_nr_n \geq 1$ will always be active, i.e., $w_nr_n = 1$.  So, in summary, we can reformulate \eqref{prb:MINLP} as the following Conic program:
  \begin{align}
    \max_{\bx,\bw, \by,\br} & \qquad \sum_{n\in \cN}q_{n} -  \sum_{n\in \cN}q_n U^c_n w_{n} \label{prb:CONIC}\tag{\sf CONIC}  \\
    \text{s.t.} & \qquad \sum_{i\in [m]} x_i \leq C \nonumber \\
     &\qquad r_n =  \sum_{i\in [m]}V_{ni}x_i + U^c_n \qquad \forall n\in \cN\nonumber\\
    &\qquad r_n w_n \geq 1 \qquad \forall n\in \cN \label{eq:cone}\\
    &\qquad \text{Routing constraints:  \eqref{eq:start-end}--\eqref{eq:subtour} } \nonumber\\
&\qquad \bx  \in \{0,1\}^m, \by \in \{0,1\}^{(m+1)\times (m+1)}\nonumber    
\end{align}

In comparison with the Conic reformulations proposed in prior works \citep{Sen2017,mehmanchi2020robust}, \eqref{prb:CONIC} does not require introducing new variables  to represent bilinear terms $x_iw_n$ and does not involve the transformation from $x_i$ to $x_i^2$ (to convert non-convex constraints to rotated cones), which typically causes weak continuous relaxations. 

\section{Cutting Plane and B\&C Approaches}\label{sec: CP and BC}

The MILP and Conic reformulations presented above would not be scalable in handling large instances, as shown in prior works and our later experiments. This is primarily due to the fact that the linearization and rotated cone transformation techniques utilized above often cause weak continuous relaxations. Moreover, the routing constraints involve numerous additional binary variables, making them expensive to handle. In this section, we present our approaches to scale up the optimization. The idea is to introduce valid cuts to approximate the nonlinear objective function and routing constraints. We then show how to incorporate such valid linear cuts into Cutting Plane and B\&C methods to achieve better scalability.

To start, let us first divide the set of zones $\cN$ into $\cL$ disjoint groups $\cD_1,\ldots,\cD_{\cL}$ such that  $\bigcup_{l\in [\cL]} \cD_l = \cN$. The objective function then can be written as: $f(\mathbf{x}) = \sum_{l\in [\cL]} \Psi_l(\mathbf{x})$, where
\[
\Psi_l(\bx) =  \sum_{n\in \cD_l} q_n -  \sum_{n\in \cD_l}\frac{q_n U^c_n}{U^c_n+  \sum_{i\in [m]} V_{ni}x_i},\:\forall l\in[\cL].
\]

Note that if $\cL = 1$, then there is only one group, and cuts will be created for the whole objective function. On the other hand, if $|\cD_l| = 1$, the cuts are created for each single ratio of the objective function. While the former setting will result in a smaller number of cuts, it often requires more iterations to converge, compared to the later setting in \cite{MaiLodi2020_OA}. So, to achieve good performance,  \cite{MaiLodi2020_OA} suggest that one should divide the set of zones $\cN$ into some smaller groups, in such ways that balance between the number of cuts created at each iteration, and the number of iterations until convergence. 

 With the above separation, we rewrite the mixed-integer nonlinear program \eqref{prb:MINLP} as
   \begin{align}
    \max_{\bx,\by} & \qquad \sum_{l\in [\cL]}\theta_l \label{prb:master-raw}\tag{\sf MINLP-2}\\
    \text{s.t.} & \qquad \sum_{i\in [m]} x_i \leq C \nonumber \\
    &\qquad \theta_l \leq \Psi_l(\bx) \qquad \forall l\in [\cL]  \label{ctr:psi}\\
    &\qquad \text{Routing constraints:  \eqref{eq:start-end}--\eqref{eq:subtour} } \nonumber\\
&\qquad \bx  \in \{0,1\}^m, \by \in \{0,1\}^{(m+1)\times (m+1)}\nonumber    
\end{align}
We will explore, in the following, three types of cuts to handle the nonlinear constraints \eqref{ctr:psi} and the routing constraints \eqref{eq:start-end}--\eqref{eq:subtour}. 

\subsection{Outer-approximation and Submodular Cuts} 

\paragraph{Outer-approximation cuts.} The outer-approximation method \citep{Duran1986} is a popular approach to solve mixed-integer nonlinear programs with convex objective functions and constraints. In the context of MCP, outer-approximation has become a state-of-the-art approach, thanks  to the concavity of the objective function \citep{Ljubic2018outer, MaiLodi2020_OA}. Specifically, in the context of the MCP-R, it can be seen that each component $\frac{-q_nU^c_n}{U^c_n + \sum_{i\in [m]} V_{ni}x_i}$ is concave in $\bx$. As a result, $\Psi_l(\bx)$ is also concave in $\bx$. The concavity then  implies that, for any solution candidate $\overline{\bx} \in \{0,1\}^m$, the following inequality holds:
\[
 \Phi_l(\bx) \leq \nabla_{\bx} \Psi_l(\overline{\bx})^\transpose (\bx-\overline{\bx}) + \Psi_l(\overline{\bx})
\]

which further implies that the following linear cuts are valid for  constraints \eqref{ctr:psi}:
 \begin{equation}
 \label{eq:OA-cuts}
 \theta_l \leq  \nabla_{\bx} \Psi_l(\overline{\bx})^\transpose (\bx-\overline{\bx}) + \Psi_l(\overline{\bx}),~~\forall l\in[\cL]     
 \end{equation}
An outer-approximation algorithm, in a cutting-plane fashion,  works as follows. We first consider a master problem, which is \eqref{prb:master-raw} but without the nonlinear constraints \eqref{ctr:psi}. At each iteration,  we solve the master problem to obtain a solution candidate  $(\overline{\btheta},\overline{\bx})$. If $(\overline{\btheta},\overline{\bx})$ is feasible to the original problem \eqref{prb:master-raw},  then we can return $\overline{\bx}$
as an optimal solution. Otherwise, outer-approximation cuts of the form \eqref{eq:OA-cuts} are added to the master problem and we move the the next iterations. It can be shown that this procedure will always terminate at an optimal solution after a finite number of iterations \citep{Duran1986}. 

\paragraph{Submodular cuts.}  Besides the concavity, it is known that the objective function $f(S)$ is submodular. The submodularity, together with the monotonicity, implies that a simple poly-time greedy algorithm can always return at least a $(1-1/e)$ approximation solution \citep{nemhauser1981maximizing}. The submodularity is also useful to derive valid cuts for a cutting plane or B\&C approach. To describe such  cuts, we first redefine $\Psi_l(\cdot)$ as a set function:
\[\forall l\in[\cL], ~S \subset [m]:\]
\[
\qquad \Psi_l(S) = \sum_{n\in \cD_l} q_n -  \sum_{n\in \cD_l}\frac{q_n U^c_n}{U^c_n+  \sum_{i\in S} V_{ni}}
\]
Then, we see that  $\Psi_l(S)$ is monotonically increasing and submodular in $S$. According to the properties of submodular functions, the following inequalities hold: 
\[
\forall l\in [\cL], S\subset [m],~ j\in [m]:
\]
\[
\qquad \Psi_l(S+j) \geq \Psi_l(S)
\]
\[
\forall l\in [\cL], S\subset S'\subset [m],~ j\in [m]\backslash S':
\]
\[
\qquad \Psi_l(S+j) - \Psi_l(S) \geq \Psi_l(S' + j) -  \Psi_l(S')
\]
where $S+j$ denotes the set $S\cup \{j\}$, for ease of notation.
Moreover, for a group $l\in [\cL]$, a set ${S}\in [m]$, and a location $k \in [m]$, Let us define
\[
\rho_{lk}(S) = \Psi_l ({S}+k) -  \Psi_l ({S})\]
\[=\sum_{n\in \cD_l} \frac{q_n V_{nk}}{(1+\sum_{j\in {S}} V_{nj})(1+\sum_{j^{'}\in {S}\cup k}V_{nj^{'}})}
\]
 The functions $\rho_{lk}(S)$ are often referred to as marginal gains, i.e.,  gains from adding an item $k$ to the set $S$. The submodular properties imply that $\rho_{lk}(S)\geq 0$ for  any $S\subset [m]$ and 
 \[
 \rho_{lk}(S) \geq \rho_{lk}(S'), ~~ \forall S\subset S' \subset [m]
 \]
i.e.,  marginal gains of adding an element $k$ diminish with the size of the set. These properties offer the following inequalities that hold for any subset $\overline{S}, S \subset [m]$ \citep{nemhauser1981maximizing}:
\begin{align}
    \Psi_{l}(S) &\leq \sum_{k\in ([m]\setminus \overline{S})\cap S}\rho_{lk}(\overline{S})-\sum_{k\in \overline{S}\backslash S}\rho_{lk}([m]-k) \nonumber\\
    &+  \Psi_l (\overline{S})  \nonumber\\
    \Psi_{l}(S) &\leq \sum_{k\in ([m]\setminus \overline{S}\cap S)}\rho_{lk}(\varnothing )-\sum_{k\in \overline{S}\backslash S}\rho_{lk}(\overline{S}-k)\nonumber\\
    &+  \Psi_l (\overline{S})  \nonumber
\end{align}
To transform the above inequalities into valid linear cuts, let us define the binary representation   of the marginal gain function as:
\[
\rho_{lk}(\bx) = \Psi_l(\bx + \bbe_k) - \Psi_l(\bx),~\forall l\in [\cL], k\in [m].
\]
where $\bbe_k$ is a vector of size $m$ with zero elements except the $k$-th element which takes a value of 1. 
The following linear cuts  can be deduced from the above submodular inequalities:
\begin{align}
    \theta_l &\leq \sum_{k\in [m]}\rho_{lk}(\overline{\bx})(1-\overline{x}_k)x_k \nonumber\\
    &-\sum_{k\in [m]}\rho_{lk}(\bbe - \bbe_k)\overline{x}_k(1-x_k) +  \Psi_l (\overline{\bx}) & \label{cons:sub-1}\\
    \theta_l &\leq \sum_{k\in [m]}\rho_{lk}(\textbf{0})\nonumber\\
    &-\sum_{\substack{k\in [m]\\\overline{x}_k=1}}\rho_{lk}(\overline{\bx}-\bbe_k) \overline{x}_k (1-x_k) +  \Psi_l (\overline{\bx}) & \label{cons:sub-2}
\end{align}
where $\bbe$ is an all-one vector of size $m$ and $\textbf{0}$ is an all-zero vector of size $m$. 

\paragraph{Cutting plane with outer-approximation or submodular cuts.} 
As far as our current knowledge goes, previous research has only employed these cuts within a B\&C procedure, where achieving convergence is straightforward. In the context of the MCP, these cuts can be utilized in a cutting plane manner, wherein cuts are sequentially generated and added to a master problem that is solved iteratively until an optimal solution is obtained. This prompts the question of whether a cutting plane method incorporating submodular cuts can yield an optimal solution, bearing in mind that when outer-approximation cuts are utilized, prior studies have demonstrated that an optimal solution can be reached after a finite number of iterations \citep{Duran1986,OA_Bonami2008algorithmic}. To explore this question, let us introduce a general concept of \textit{valid cuts}, which allows us to unify both  outer-approximation and submodular cuts presented above within a   cohesive framework.

\begin{definition}[Valid cut]\label{def:valid-cut}
For any candidate solution $\overline{\bx}$, a  linear cut $\theta_l \leq (\bap_{\overline{\bx}}^l)^\transpose \bx +\beta^l_{\overline{\bx}}$ (the parameters $(\bap_{\overline{\bx}}^l)$ and $\beta^l_{\overline{\bx}}$ are dependent of $\overline{\bx}$) is said to be a valid cut if 
\[
 (\bap_{\overline{\bx}}^l)^\transpose {\bx} +\beta^l_{\overline{\bx}} \geq \Psi_l(\bx), ~\forall \bx\in \{0,1\}^m
 \]
 \[
 (\bap_{\overline{\bx}}^l)^\transpose \overline{\bx} +\beta^l_{\overline{\bx}} = \Psi_l(\overline{\bx})
\]
\end{definition}

In other words,  a linear cut  $(\bap_{\overline{\bx}}^l)^\transpose \bx +\beta^l_{\overline{\bx}}$ is valid if the linear function $(\bap_{\overline{\bx}}^l)^\transpose \bx +\beta^l_{\overline{\bx}}$ lies above the concave function $\Psi_l(\bx)$ and intersect with it at the point $\overline{\bx}$. Within the binary domain, it can be seen that there would be infinitely many valid cuts crossing at any specific point $\overline{\bx} \in \cX$, and the outer-approximation and sub-modular cuts given in  \eqref{eq:OA-cuts}, \eqref{cons:sub-1} and  \eqref{cons:sub-2} are indeed valid ones. Proposition \eqref{prp:milp-valid-cuts} below states our first result saying that the mixed-integer nonlinear program \eqref{prb:MINLP} is equivalent to a MILP where the nonlinear constraints {\eqref{ctr:psi}} are replaced by valid cuts generated at every binary point in the feasible set $\cX$. 
\begin{proposition}\label{prp:milp-valid-cuts}
    The  mixed-integer nonlinear program \eqref{prb:master-raw} is equivalent to the following MILP:
       \begin{align}
    \max_{\bx,\by} & \qquad \sum_{l\in [\cL]}\theta_l \label{prp:MNLP-valid}\tag{\sf MILP*}\\
    \text{s.t.} & \qquad \sum_{i\in [m]} x_i \leq C \nonumber \\
    &\qquad \theta_l \leq (\bap_{\overline{\bx}}^l)^\transpose \bx +\beta^l_{\overline{\bx}} \qquad \forall \overline{\bx}\in \cX \label{eq:valid cuts} \\
    &\qquad \text{Routing constraints:  \eqref{eq:start-end}--\eqref{eq:subtour} } \nonumber\\
&\qquad \bx  \in \{0,1\}^m, \by \in \{0,1\}^{(m+1)\times (m+1)}\nonumber    
\end{align}
\text{where linear constraints \eqref{eq:valid cuts} are valid cuts.} 
\end{proposition}

\begin{proofs}
We will validate the equivalence by proving that the feasible sets of \eqref{prb:master-raw} and \eqref{prp:MNLP-valid} are identical. To this end, let $\Omega$ be the feasible set of \eqref{prb:master-raw}  and $\Omega'$ be the feasible set of \eqref{prp:MNLP-valid}. We now see that for any $(\btheta,\bx,\by) \in \Omega$, we have $\theta_l \leq  \Psi_l(\bx)$. Since all the valid cuts lie above $\Psi_l(\bx)$, the following inequalities hold
\[
\theta_l \leq  \Psi_l(\bx) \leq (\bap^l_{\overline{\bx}})^\transpose \bx +\beta^l_{\overline{\bx}},~\forall \overline{\bx}\in \cX
\]
implying that $(\btheta,\bx,\by) \in \Omega'$. On the other hand, if $(\btheta,\bx,\by) \in \Omega'$, we also have
\begin{align*}
    \theta_l &\leq \min_{\overline{\bx}} \{(\bap^l_{\overline{\bx}})^l \bx +\beta^l_{\overline{\bx}}\} \\
    &\leq (\bap^l_{{\bx}})^\transpose \bx +\beta^l_{{\bx}}\\
    &= \Psi_l(\bx)
\end{align*}
which implies that $(\btheta,\bx,\by)$ is also feasible to \eqref{prb:master-raw}. So, we can conclude that $\Omega$  and $\Omega'$ are identical, as desired. 
\end{proofs}

Here we note that in the seminal work of \cite{nemhauser1981maximizing}, it is demonstrated that the submodular maximization problem can be reformulated equivalently as a MILP whose constraints are submodular cuts generated at every point within the binary domain. Proposition \ref{prp:milp-valid-cuts} generalizes this result by showing that the binary nonlinear optimization problem can be reformulated as a MILP with ``\textit{valid cuts}'' generated at every binary point in the feasible set. Such valid cuts can be outer-approximation or submodular cuts, or any linear cuts satisfying Definition \ref{def:valid-cut}.

We now explore a general question asking whether a cutting plane procedure with \textit{valid cuts} would converge to an optimal solution after a finite number of iterations. To formally state the result, let us describe below the cutting plane procedure with valid cuts. 

\begin{framed}
\textbf{{Cutting plane with valid cuts:}} \textit{Define a \textbf{master problem} as \eqref{prb:master-raw} but without  the nonlinear constraints \eqref{ctr:psi}. At each iteration:
\begin{itemize}
    \item Solve the master problem  to obtain a solution candidate $(\overline{\btheta},\overline{\bx},\overline{\by})$
    \item If $\theta_l\leq \Psi_l(\overline{\bx})$ for all $l\in [\cL]$, then return $\overline{\bx}$ as an optimal solution and stop the procedure
    \item Add valid cuts $(\bap_{\overline{\bx}}^l)^\transpose \bx +\beta^l_{\overline{\bx}}$, $l\in [\cL]$ to the master problem.  
\end{itemize}}
\end{framed}

The following theorem states that the above ``\textit{Cutting Plane with valid cuts}'' procedure will always return an optimal solution to \eqref{prb:master-raw} after a finite number of iterations. 

\begin{theorem}\label{th:1}
If  \eqref{prb:master-raw} has at least one solution, then the cutting plane with valid cuts described above always returns an optimal solution after a finite number of iterations.
\end{theorem}
\begin{proofs}
    Let us denote by $\cK$ the set of solution candidates $(\btheta,\bx,\by)$ collected during the cutting plane procedure. We will prove that if solving the master problem gives a solution candidate $(\btheta^*,\bx^*,\by^*)$ such that $\bx^*$  already appears in $\cK$, then $(\btheta^*,\bx^*,\by^*)$ is optimal for \eqref{prb:master-raw}. 
  To achieve this, we first note that the master problem contains cuts generated by points in $\cK$, thus 
  \[
  \theta^*_l \leq (\bap^l_{\bx})^l \bx^* +\beta^l_{\bx};~\forall l\in [\cL], \bx \in \cK
 \]
Since $\bx^*$ is already included in $\cK$, we have
\[
    \theta^*_l \leq (\bap^l_{\bx^*})^l \bx^* +\beta^l_{\bx^*} = \Psi_l(\bx^*);~\forall l\in [\cL]
\]
  which implies that $(\btheta^*,\bx^*,\by^*)$ is feasible to \eqref{prb:master-raw}. Moreover, we see that the feasible set of the master problem contains that of \eqref{prp:MNLP-valid}. The equivalence stated in Proposition \ref{prp:milp-valid-cuts} further implies that the feasible set of the master problem contains the feasible set of \eqref{prb:master-raw}.  We now  consolidate our observations as follows:
\begin{itemize}
    \item $(\btheta^*,\bx^*,\by^*)$ is optimal for \textit{master problem}
    \item $(\btheta^*,\bx^*,\by^*)$ is feasible to \eqref{prb:master-raw}
    \item The feasible set of the master problem contains the feasible set of \eqref{prb:master-raw}.
\end{itemize}

The three observations above are sufficient to see that $(\btheta^*,\bx^*,\by^*)$ is optimal to \eqref{prb:master-raw}.  The feasibility of $(\btheta^*,\bx^*,\by^*)$ to \eqref{prb:master-raw} also implies that  the  stopping conditions are met and the cutting plane will terminate and return this optimal solution.

To demonstrate the termination of the cutting plane method, it is essential to see that the feasible set $\cX$ is finite. Therefore, after a finite number of iterations, the cutting plane procedure will either stop or yield a solution $\bx$ that is already present in $\cK$. Based on the analyses above, we can assert that at this point, the candidate solution is optimal, and the algorithm will terminate.
\end{proofs}

\subsection{Sub-tour Elimination Cuts}

In this section, we introduce sub-tour elimination cuts specifically designed to effectively handle the routing constraints. The idea here is to substitute the routing constraints from \eqref{eq:start-end} to \eqref{eq:subtour}, with more manageable relaxed constraints. Although these relaxed constraints are easier to work with, they would result in infeasible tour solutions. We then utilize the  sub-tour elimination cuts to eliminate these infeasible solutions from the feasible set.

Let us start by  not including the location $m+1$ (i.e., the last \textit{depot}) in the formulation, i.e., we only consider location  $0$ as the depot, and the set $\{1, 2, ..., m\}$ also represents the available locations for locating new facilities.
The binary variables $x_i,\; \forall i \in [m] \cup {0}$ are the same as in previous sections, but the binary variables $y_{ij}$ are defined differently to reduce the number of binary variables and constraints. That is, $y_{ij} = 1, \; i,j \in V\cup \{0\}, \; i < j$ if and only if there exists visit between $i$ and $j$. With these new variable definition, we formulate a master problem  as follows:
\begin{align}
    \max_{\btheta,\bx,\by}  &\qquad \sum_{l\in [\cL]}\theta_l \label{prb:master-subtour} \tag{\sf MINLP-3}\\
      \text{s.t.} & \qquad \sum_{i\in [m]} x_i \leq C \nonumber \\
    &\qquad \theta_l \leq \Psi_l(\bx) \qquad \forall l\in [\cL]  \nonumber\\
    &\qquad  \sum_{i = 0}^{m} \sum_{j = i+1}^{m} t_{ij}y_{ij} \leq T_{max} \label{eq:Tmax} \\
    &\qquad  \sum_{i < j} y_{ij} + \sum_{j < k} y_{jk} = 2x_j \qquad \forall i,j \in V \label{eq:in-out} \\
    &\qquad  x_0 = 1 \label{eq:x-0} \\
    &\qquad  \bx \in \{0,1\}^{m+1}, \; \by \in \{0,1\}^{(m+1)^2} \nonumber
\end{align}
Note that in \eqref{prb:master-subtour} we keep the nonlinear constraints $\delta_l\leq \Psi_l(\bx),~ \forall l\in [\cL],$ for the sake of completeness. These constraints will be replaced by outer-approximation or sub-modular cuts during the cutting plane or B\&C procedure, as described above. Solving \eqref{prb:master-subtour} may give invalid routing solutions that contain sub-tours. To address this issue, the Branch-and-cut algorithm for the OP is first time introduced by \cite{Fischetti}. The algorithm uses several families of valid inequalities and conditional cuts to deduce the OP solution space. In our research, we adopt two types of constraints inspired by \cite{Nguyen2021AnEB} for our B\&C framework. These constraints are specifically designed to prevent the formation of sub-tours and ensure a valid routing solution. The first type of constraints involves forcing the presence of at least two edges between any set $S$ and $V \backslash S$, for any subset $S$ of $V$ such that $S$ contains at least one selected location. The constraints can be formulated as follows:
\[
    \sum_{\substack{i\in S \\j\in V \backslash S \\ i< j}}y_{ij} + \sum_{\substack{i\in V \backslash S\\ j\in S \\ i< j}}y_{ij} \geq 2 x_k, \; \forall k\in S, S \subseteq V, |S| \geq 3 \label{eq:SEC-1}
\]
Additionally, the second type of constraints  is a sum of the first type constraints, which is formulated as follows:
\[
    |S|\sum_{\substack{i,j \in S\\i < j}}y_{ij} \leq (|S| - 1) \sum_{k\in S} x_k, \; \forall |S| \geq 3, S \subseteq V \label{eq:SEC-2}
\]
Similar to the outer-approximation or submodular cuts mentioned above, these sub-tour elimination constraints  can be generated and added to the master problem \eqref{prb:master-subtour}, in a cutting plane or B\&C fashion,  for any invalid routing solution $\overline{\by}$.   By incorporating these two types of constraints, we ensure that infeasible routing solutions  will be eliminated from the feasible set of \eqref{prb:master-subtour}  during the optimization process.

\subsection{Algorithms}
\label{Cut-Algo}
We described our cutting plane and B\&C algorithm based on the  three types of cuts presented above   (i.e., outer-approximation, submodular and sub-tour elimination cuts). 

\paragraph{Nested Cutting Plane.}
Our cutting plane  method is based on the following master problem, where the nonlinear constraints and part of the routing constraints in \eqref{prb:master-raw}  are replaced by outer-approximation, submodular and sub-tour elimination cuts:
\begin{align}
    \max_{\btheta,\bx,\by}  &\qquad \sum_{l\in [\cL]}\theta_l \label{prb:master-all} \tag{\sf CP-Master}\\
      \text{s.t.} & \qquad \sum_{i\in [m]} x_i \leq C \nonumber \\
    &\qquad  \sum_{i = 0}^{m} \sum_{j = i+1}^{m} t_{ij}y_{ij} \leq T_{max} \nonumber \\
    &\qquad  \sum_{i < j} y_{ij} + \sum_{j < k} y_{jk} = 2x_j \qquad \qquad \forall i,j \in V \nonumber \\
    &\qquad  x_0 = 1 \nonumber \\
   &\qquad  \textit{[Sub-tour elimination cuts]} \nonumber \\
   &\qquad  \textit{[Outer-approximation and submodular cuts]} \nonumber \\
    &\qquad  \bx \in \{0,1\}^{m+1}, \; \by \in \{0,1\}^{(m+1)^2} \nonumber
\end{align}
Our nested cutting plane algorithm involves an iterative procedure where cuts are iteratively added to the master problem \eqref{prb:master-all} in a nested fashion. At each outer iteration, we solve the current master model and obtain a solution $(\overline{\btheta},\overline{\bx},\overline{\by})$. In each inner iteration, if the current solution contains sub-tours, indicating that the routing solution $\overline{\by}$ is not valid, we introduce sub-tour elimination cuts to the master problem and re-solve the updated master problem to obtain a valid solution. If no sub-tours are detected in the found solution (indicating feasibility), and the current solution is feasible (under a given approximation threshold) for the MCP-R in \eqref{prb:master-raw}, then we stop the algorithm and return the current solution. Otherwise, we add outer-approximation cuts and submodular cuts to the master problem and proceed to the next outer loop. We present our nested cutting plane algorithm in Algorithm \ref{algo:CP}  below.

It can be shown that this nested cutting plane procedure always terminates at an optimal solution (assuming there is at least one) after a finite number of inner and outer iterations.

\begin{theorem}[Convergence of the Nested Cutting Plane]
   Assuming that the MCP-R problem has at least one feasible solution, then Algorithm \ref{algo:CP} always terminates after a finite number of outer and inner steps,  and returns an optimal solution. In particular,   let $\cK = \{(\btheta_k,\bx_k, \by_k);~ k =1,\ldots\}$ be  the set of solution candidates collected from solving the master problem . If at Line 5,  solving the master problem \eqref{prb:master-all} yields a solution $(\overline{\btheta},\overline{\bx},\overline{\by})$ such that $\overline{\by}$ forms a valid tour, and there is a solution $(\btheta_k,\bx_k, \by_k) \in \cK$ such that $\bx_k = \overline{\bx}$, then $\overline{\bx}$ is an optimal solution to the MCP-R problem and the stopping conditions satisfy. . 
\end{theorem}
\begin{proofs}
  We first show that for each iteration of the outer loop, the inner loop will always terminate after a finite number of iterations. To see why it should be the case, let us consider an outer iteration with outer-approximation an submodular cuts generated from solutions in $\cK$. To facilitate our later exposition, let $\cH$ be the set of all the parameter pairs $(\bap^l,\beta^l)$ such that $\theta_l \leq (\bap^l)^\transpose \bx +\beta^l$ are outer-approximation or submodular cuts generated from points in $\cK$.
 We have the following observations:
  \begin{itemize}
      \item [(i)] The number possible  $(\bx,\by)$ solutions (where $\by$ does not necessarily form  a valid tour) is finite -- this is obviously seen as $\bx$ and $\by$ are binary variable of finite sizes.
      \item [(ii)] After each iteration of  the inner loop,  an invalid  solution $(\bx,\by)$ will be removed from the feasible  set of the master problem.
      \item[(iii)] Within an outer iteration, for any two solution candidates $(\btheta_1,\bx_1,\by_1)$ and $(\btheta_2,\bx_2,\by_2)$ obtained from solving the master problem, if $(\bx_1,\by_1)  = (\bx_2,\by_2)$, then  $\btheta_1 = \btheta_2$. This is because, within the same outer iteration, there are no new  outer-approximation and submodular cuts  being added to the master problem, so the sets $\cK$  and  $\cH$ remain unchanged. Since the sub-tour cuts do not affect the variables $\btheta$, the master problem will find the maximum achievable $\btheta$ to satisfy the outer-approximation and submodular cuts. This implies:
      \begin{align}\small
          \btheta_1 &= \text{argmax}_{\btheta}\left\{\sum_{l\in [\cL]}\theta \Big|~ \begin{cases}
              \theta_l\leq (\bap^l)^\transpose \bx_1 +\beta^l\\ 
          \forall (\bap^l,\beta^l)\in \cH
          \end{cases}\right\} \nonumber\\
                \btheta_2 &= \text{argmax}_{\btheta}\left\{\sum_{l\in [\cL]}\theta \Big|~ \begin{cases}
                    \theta_l\leq (\bap^l)^\transpose \bx_2 +\beta^l \\ 
          \forall (\bap^l,\beta^l)\in \cH
                \end{cases}\right\} \nonumber
      \end{align}
      Since $\bx_1 = \bx_2$, we should have $\btheta_1 = \btheta_2$.
  \end{itemize}
Under the assumption that the MCP-R has at least a feasible solution, the above observations indicate that the inner loop always finitely terminate at a feasible solution.

We now discuss the main result. Similar to the proof of Theorem \ref{th:1}, we see that if $(\overline{\btheta},\overline{\bx},\overline{\by} )$ is a solution given by the master problem containing valid cuts generated by points in $\cK$, we should have 
  \[
  \overline{\theta}_l \leq (\bap^l_{\bx})^l \overline{\bx} +\beta^l_{\bx};~\forall l\in [\cL], \bx \in \cK
 \]
So, if we see that $\overline{\bx}$ is already included in $\cK$, we have
\[
    \overline{\theta}_l \leq (\bap^l_{\overline{\bx}})^l \overline{\bx} +\beta^l_{\overline{\bx}} = \Psi_l(\overline{\bx});~\forall l\in [\cL]
\]
which implies that $(\overline{\btheta},\overline{\bx},\overline{\by})$ is feasible to \eqref{prb:master-raw}. Moreover,  the feasible set of the master problem \eqref{prb:master-all} contains that of \eqref{prb:master-raw}. So, if we let $(\btheta^*,\bx^*,\by^*)$ be an optimal solution to \eqref{prb:master-raw},   we will have
$\sum_{l\in [\cL]}\overline{\theta}_l \ge \sum_{l\in [\cL]}{\theta}^*_l $. Combine this with the remark that $(\overline{\btheta},\overline{\bx},\overline{\by})$ is feasible to \eqref{prb:master-raw}, it is clear that $(\overline{\btheta},\overline{\bx},\overline{\by})$  is optimal to \eqref{prb:master-raw} and the algorithm should stop. To understand why this will always happen after a finite number of outer iterations, we observe that there is only a finite number of solutions that can be added to $\cK$. This is because there are only a finite number of possible values for $(\bx,\by)$, and for each pair $(\bx,\by)$, Observation $(iii)$ above tells us that there is only a unique $\btheta$ produced by solving the master problem. 
  \end{proofs}
\begin{algorithm}[htb]
    \caption{Nested Cutting Plane (\textbf{NCP})} 
    \label{algo:CP}
    \SetKwRepeat{Do}{do}{until}
Set a small $\epsilon>0$ as the optimal gap, $\cK = \emptyset$ \\
Solve the master problem \eqref{prb:master-all} to get a solution candidate  $(\overline{\btheta},\overline{\bx},\overline{\by} )$\\
\comments{Outer loop}\\
\Do{$\overline{\theta}_l \leq \Psi_l(\overline{\bx}) + \epsilon,~\forall l\in [\cL]$, and $\overline{\by}$ forms a valid tour.
    }
    {
    \eIf{ ($\overline{\by}$ forms a valid tour) }
    {
    $\cK \leftarrow \cK \cup (\overline{\btheta},\overline{\bx},\overline{\by} )$ \comments{Add the valid solution to the solution pool} 
    }
    {
    \comments{Inner loop} \\
    \Do {$\overline{\by}$ forms a valid tour.}
    {
      Add sub-tour elimination cuts  for $\overline{\by}$ to the master problem \eqref{prb:master-all}\\
      Solve \eqref{prb:master-all}  to get a new solution candidate $(\overline{\btheta},\overline{\bx},\overline{\by} )$ 
    }
        }
  Add outer-approximation and submodular cuts  for $(\overline{\btheta},\overline{\bx},\overline{\by} )$  to \eqref{prb:master-all}\\
  Solve \eqref{prb:master-all} to get a new solution candidate $(\overline{\btheta},\overline{\bx},\overline{\by} )$ 
  
    }
    Return $(\overline{\by},\overline{\bx})$ as an optimal solution.
\end{algorithm}
\paragraph{Nested B\&C:} Alternatively, a B\&C approach can be developed where outer-approximation, submodular and sub-tour elimination cuts are added in a similar nested fashion. We can do this using the lazy-cut callback and user-cut callback procedure within a MILP solver (e.g. CPLEX). At each node of the branch and bound tree, similar to the cutting plane approach, the MILP solver checks if the current solution contains sub-tours, then it adds sub-tour elimination cuts to eliminate the sub-tours. Conversely, if no sub-tours are present, the solver adds outer-approximation and submodular cuts if the stopping conditions are not satisfied. We note that only sub-tour elimination cuts are implemented within the user-cut callback.

\section{Metaheuristic}\label{sec:Heuristic}
Our exact methods developed above would encounter difficulties when dealing with large-scale instances, especially when the number of locations significantly increases. To address this challenge, we design a metaheuristic approach which is based on the Iterated Local Search (ILS) procedure by \cite{Gunawan2015}. Algorithm \ref{alg:ils} describes the pseudo-code of our ILS. It begins with \texttt{Construction} operator, a greedy construction heuristic to generate an initial feasible solution, which is then further enhanced by an integration of three main operators: (i) \texttt{LocalSearch}, which improves the solution initially obtained; (ii) \texttt{Perturb}, where a new starter point is generated through a perturbation of the solution provided by the \texttt{LocalSearch}, (iii) \texttt{AcceptanceCriterion}, that determines from which solution the search should continue. Additionally, we also propose a technique based on the Taylor approximation to quickly estimate the objective value of the solution obtained after each move of the local search operators, thus speeding up the algorithm in overall while keeping its quality.

\begin{algorithm}[htb]
\caption{Iterated Local Search (ILS)}\label{alg:ils}
$S_0 \gets \texttt{Construction}$\\
$S_0 \gets \texttt{LocalSearch}(S_0, \cU,\cV)$\\
$S \gets S_0$; 
$nb_{imp} \gets 1$\\
\While{\textit{Stopping criterion is not meet}}{
    $S_0 \gets \texttt{Perturb}(S_0, \cU,\cV)$\\
    $S_0 \gets \texttt{LocalSearch}(S_0, \cU,\cV)$\\
  \eIf{$S_0$ better than $S$}{
    $S \gets S_0$;
    $nb_{imp} \gets 1$\;
  }{
    $nb_{imp} \gets nb_{imp} + 1$\\
  \If{$nb_{imp} \mod threshold = 0$}{
    $S_0 \gets S$\;
    }
  }
}
\KwResult{$S$}
\end{algorithm}

\SetKwComment{Comment}{/* }{ */}

\subsection{Solution Construction}
In the greedy construction phase, we aim to build an initial solution to start the procedure. The idea is to keep adding vertices to a route until there are no more vertices that can be inserted successfully. If the total time required to visit every vertex in the route after the insertion $\overline{T}$ does not exceed the time budget $T_{max}$, the vertex insertion is feasible. At each iteration of the construction process, all candidate vertices, which can be inserted into the route in at least a feasible position, are examined. Then the candidate vertex contributing the highest utility to the route will be added to the position that increases the lowest time. Let $\cV$ and $\cU$ be the sets of unvisited and visited vertices, respectively. The detail of the initial solution construction is presented in Algorithm \ref{alg:construction}.

The result of the \textbf{for} loop from Line 7 to Line 10 of Algorithm \ref{alg:construction} is the vertex $i$ which contributes the most utilities to the objective value. In Line 9, the function $Obj^I(S_0,v)$ calculates the objective value of the solution after temporarily inserting vertex $v$ into $S_0$. After finding the best vertex $i$, a procedure to decide the best position for insertion is performed in Line 12. Finally, the objective value of current solution $S_0$ is computed by function $Obj(S_0)$ and the total traveling time $\overline{T}$ is updated. It is worth noting that if the values of $\sum_{i \in [m]}V_{ni}x_i$ for each $n \in \cN$ are calculated and stored in a list, the complexity of the function $Obj(S_0)$ and $Obj^I(S_0,v)$ is just $\mathcal{O}(|\cN|)$ instead of $\mathcal{O}(m|\cN|)$ as calculating directly the objective value from the solution $S_0$.

\begin{algorithm}[htb]
\caption{\texttt{Construction}}\label{alg:construction}
    Initialize: $\cU \gets \emptyset$; $\cV \gets [m]$; $Obj(S_0) \gets 0$; $S_0 \gets \{0,0\}; \overline{T} \gets 0$ \\
    \While{$\cV \neq \emptyset$ \texttt{ and } $\overline{T} < T_{max}$}{
        Declare: $i, pos, inter \gets 0$;\\
        \comments{$i$ presents the best vertex for inserting to the route}\\
        \comments{$pos$ presents the best position for inserting $i$ to the route}\\
        \comments{$inter$ is an intermediate variable used to find the best vertex to insert}\\
        \For{$v \in \cV$}{
            \If {$\exists$ a feasible position in $S_0$ to insert $v$} {
                \If {$Obj^I(S_0,v) > inter$}{
                    $i \gets v$ and $inter \gets Obj^I(S_0,v)$
                }
            }
        }
        \If{$\exists i$}{
            Find $pos$\\
            Update solution $S_0$ (by inserting $i$ to position $pos$), $\cU$, and $\cV$\\
            Calculate $Obj(S_0)$ and $\overline{T}$
        }
        \Else{
            break
        }
    }
    \KwResult{$S_0$}
\end{algorithm}

\subsection{Local Search}
\begin{figure}[!h]
    \centering
    \subfloat[\centering $Replace_1()$]{{\includegraphics[width=0.4\textwidth]{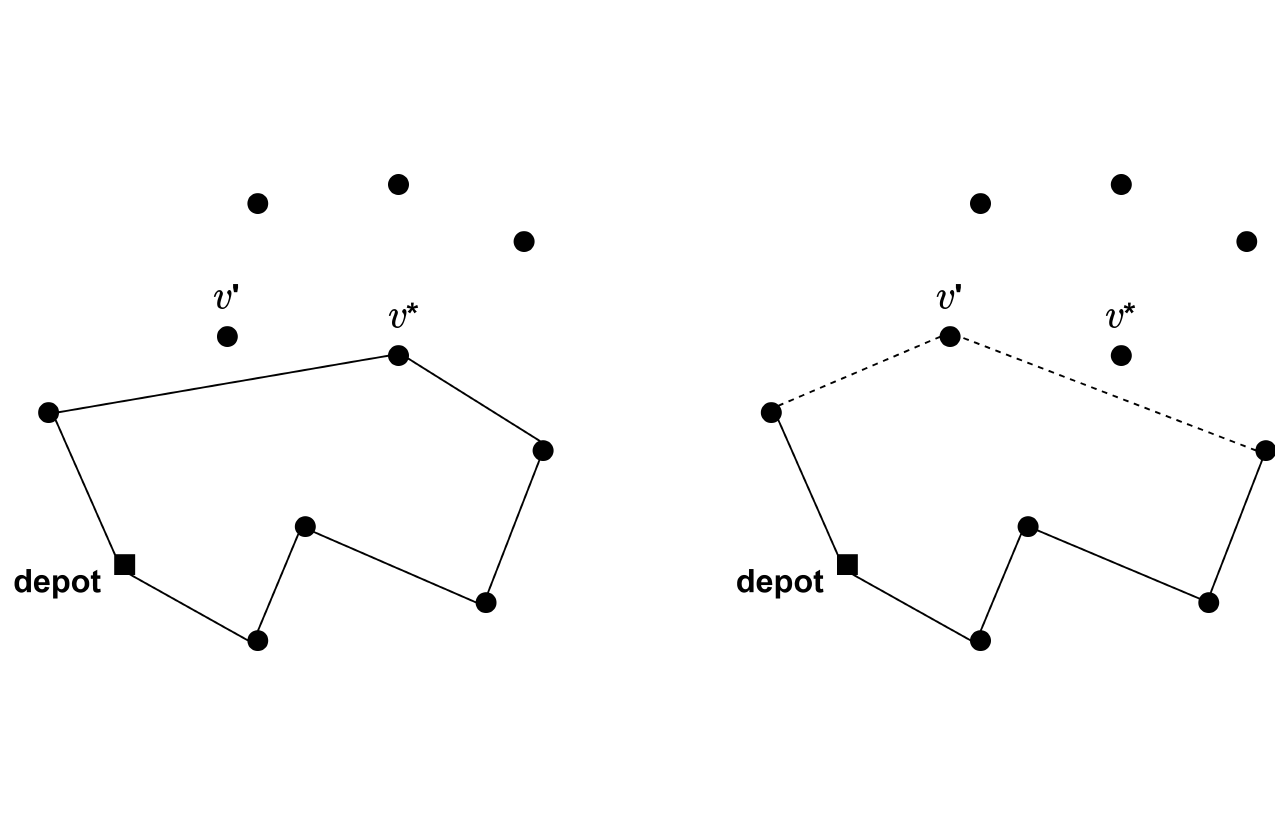} }}
    \qquad \qquad \qquad \qquad
    \subfloat[\centering $Replace_2()$]{{\includegraphics[width=0.4\textwidth]{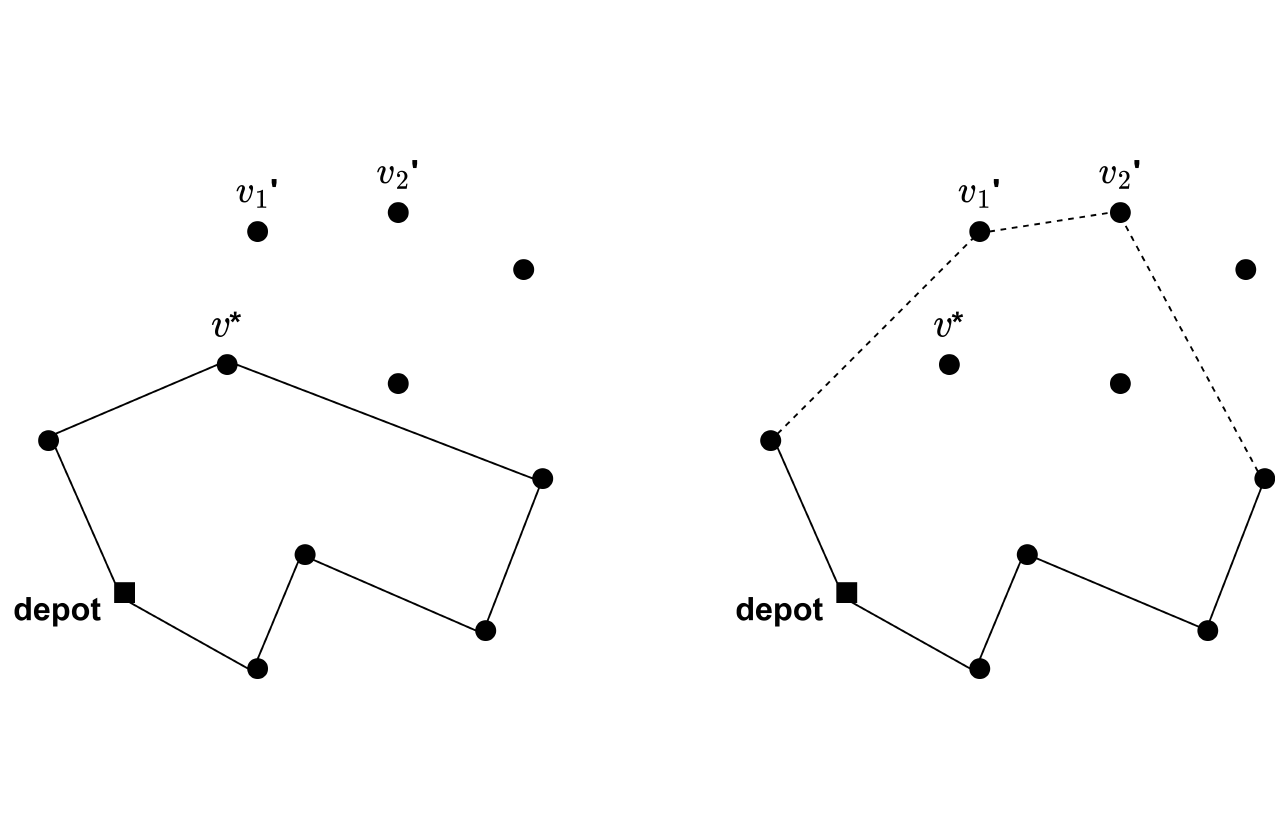} }}
    \caption{\texttt{Replace} operators used to improve the solution}
    \label{fig:replace}
\end{figure}

The basic idea of the \texttt{LocalSearch} operator is to make some small changes to a given solution in the hope of finding a new better solution. We first use two classical TSP local searches: \textit{2-Opt} \citep{2opt} and \textit{Or-Opt} \citep{OrOpt} to shorten the length of the route without removing any vertices. As such, we can have more space in the time budget to insert more unvisited vertices. We then apply two \textit{Replace} operators which try to replace one visited vertex $v^* \in \cU$ with one unvisited vertex $v'$ or one pair of unvisited vertices $\{v_1',v_2'\}$ $(v', v_1', v_2' \in \cV)$, resulting two operators: $Replace_1$ and $Replace_2$, respectively. In the former, each vertex $v^* \in \cU$ in the route is examined whether a unvisited vertex $v' \in \cV$ can replace it in its position (see Figure \ref{fig:replace}.a) by a function called  $Obj^R_1(S_0, v^*, v')$. In the latter, the change of objective value is the same when replacing a visited vertex $v^*$ with edge $(v_1',v_2')$ or reverse edge $(v_2',v_1')$. Therefore, for these two edges, we only consider the one that leads to a shorter route in terms of traveling time (see Figure \ref{fig:replace}.b). If the resulting traveling time does not exceed the time budget, the candidate edge is examined by the function $Obj^R_2(S_0, v^*, v_1', v_2')$. Once vertex $v^*$ is successfully replaced, these operations continue with the next visited vertices until the objective value cannot be improved. After replacing all potential vertices in the route with better ones, \textit{2-Opt} and \textit{Or-Opt} operators are called again in the Local Search procedure to rearrange the order of visited vertices, shortening the route. Finally, an insertion operator, which inherits the idea of the operator \texttt{Construction}, is applied to add new potential unvistited vertices.

\subsection{Perturbation}
The role of the permutation operator is to create the diversity for the search, allowing it to escape from local optima. We first remove a part of the current solution using removal operators then rebuild it to create a new solution using an insertion heuristic. Three operators \textit{Random Removal, Historical Removal}, and \textit{String Removal} are implemented in \texttt{Pertub()} function and randomly selected in each iteration. The \textit{Random Removal} randomly selects and removes $r$ vertices while keeping the order of the remaining vertices in the route. The \textit{Historical Removal} first creates a list containing removal frequencies of all vertices in ascending order. We then search for $r$ vertices in the route that have the lowest values of removal frequency, remove them subsequently, and keep the order of the other vertices. The \textit{String Removal} is used to remove $r$ consecutive vertices in the route. In these operators, the number of removal vertices $r$ is an important parameter that need to be carefully selected. By preliminary experiments, we decide to take a random integer in range $[l/3, l/2]$ for $r$. Here, $l$ is the number of vertices except the depot in the current route. After removing $r$ vertices by one of three removal operators, the route is re-constructed by an insertion operator, which extends the idea of function \texttt{Construction()}. However, instead of choosing the best vertex to insert into the route at each step, we randomly choose one of the three best vertex candidates to enhance the diversity.

\subsection{Acceptance Criterion}

In the \textbf{if-else} condition starting from Line 7 of Algorithm \ref{alg:ils}, a new local optimum solution is always accepted to become the initial solution for the next iteration of our ILS. However, if there is no improvement of $S$ found for each certain number of iterations, parameter $threshold = 10$ in our algorithm, the current best-found solution is considered as the starting point of the next search. This strategy is inspired from the work of \cite{Gunawan2015}.

\subsection{Speeding Up via Taylor Approximation}

The running time of our ILS algorithm heavily depends on the
the complexity of functions $Obj()$, $Obj^I()$, $Obj^R_1()$ and $Obj^R_2()$ which is $\mathcal{O}(|\cN|)$, where $\cN$ is the set of customer zones. To speed up our algorithm, particularly when the number of customer zones increases, we propose a technique to reduce the cost of computing the objective functions via Taylor approximation.

Our idea is to approximate the nonlinear objective function via Taylor approximation and use it to quickly evaluate the insertion and replacement operations, instead of recalculating the exact values. For each feasible solution $S$ satisfying the routing constraints, let $\bx^S$ be the binary representation of $S$, i.e., $\bx^S$ is a vector of size $m$ such that $x^S_i = 1$ if $i\in S$  and $x^S_i=0$ otherwise. The objective function of solution $S$ can be estimated via the objective value of another solution $S_0$ by the first-order Taylor polynomial approximation as follows:
\begin{align}
\label{eq:taylor}
   &f(S)\approx \widehat{f}(S) = f(S_0) \nonumber \\
    &+ \sum_{i \in [m]} \left[\left(\sum_{n\in \cN} \frac{q_n V_{ni} U^c_n}{(\sum_{j \in [m]} x^{S_0}_j V_{nj} + U^c_n)^2}\right)(x^S_i-x^{S_0}_i)\right]
\end{align}
We now can use $\widehat{f}(S)$ as an approximation of $f(S)$ to perform insertion or replacement operations, noting that $\widehat{f}(S)$ has a linear structure and will be fast to compute. Specifically, we store values $$\cF^{S_0}_i = \sum_{n \in \cN} \frac{q_nV_{ni}U^c_n}{(\sum_{j \in [m]} x^{S_0}_jV_{nj} + U^c_n)^2},~~\forall i \in [m]$$ in a list for later use because the second term of $\widehat{f}(S)$ presents the change of the approximation when an \textit{Insertion} or a \textit{Replace} operator reconstructs the route. Any insertion into solution $S_0$ to create solution $S$, for example, inserting $v$ into $S_0$, can be represented by setting $x^{S_0}_v = 0$ to $x^S_v = 1$, then the value of $\widehat{f}(S)$ increases by $\cF^{S_0}_v$. For any replacement $i \in S_0$ by $j$, the values of $x^S_{j}$ and $x^{S_0}_{j}$ are set to 1 and 0, while the values of $x^S_{i}$ and $x^{S_0}_{i}$ are set to 0 and 1. Thus, the change in $\widehat{f}(S)$ is equal to $\mathcal{F}^{S_0}_j(1-0) + \mathcal{F}^{S_0}_i(0-1) = \mathcal{F}^{S_0}_j - \mathcal{F}^{S_0}_i$.

In the \texttt{Construction} operator and related insertion operators, we search for the maximum values of $\cF^{S_0}_i ~(i \in [m])$, insert $i$ to the route, recalculate the value of the objective using $Obj()$, and update the value of $\mathcal{F}^{S_0}_i$ in the list. For the $Replace_1()$ function, we compare $\mathcal{F}^{S_0}_i$ $(i \in \cV)$ with $\mathcal{F}^{S_0}_j$ $(j \in \cU)$ to find the pair of vertices with the maximum value $\mathcal{F}^{S_0}_i - \mathcal{F}^{S_0}_j$. Assume that $(h, k)$ is the pair indexes that maximizes $\mathcal{F}^{S_0}_i - \mathcal{F}^{S_0}_j$. To ensure that the approximation does not 
reduce the real-valued objective, we use a function called $Obj^R_1(S_0, k, h)$ to check again the objective value whether to replace $k$ by $h$ or not. When all the conditions:\\
\begin{align}\small
\label{eq:replace11}
    \begin{cases}
        0 < \mathbf{max}\{\mathcal{F}^{S_0}_i - \mathcal{F}^{S_0}_j|~ i\in \cV, j\in \cU\} = \mathcal{F}^{S_0}_{h} - \mathcal{F}^{S_0}_{k}\\
        Obj^R_1(S_0, k, h) > Obj(S_0)\\
        k =  at(p)\\
        T + t_{at(p-1)h} + t_{at(p+1)h} - t_{at(p-1)k} \\
        ~~~~~~~- t_{at(p+1)k} \leq T_{max}
    \end{cases}
\end{align}
are satisfied
the replacement occurs and all related values are updated. Similarly, if\\
\begin{align}
\label{eq:replace21}
    \begin{cases}
        0 < \mathbf{max}_{k,i\in \cV, j\in \cU}\{\mathcal{F}^{S_0}_i + \mathcal{F}^{S_0}_{k} ) - \mathcal{F}^{S_0}_j| |~ k \neq i\} \\
        ~~~~~~~= \mathcal{F}^{S_0}_{i'} + \mathcal{F}^{S_0}_{k'} - \mathcal{F}^{S_0}_{j'}\\
        Obj^R_2(S_0, j', i', k') > Obj(S_0)\\
        j' =  at(p)\\
        t_{at(p-1)i'} + t_{at(p+1)k'} < t_{at(p-1)k'} + t_{at(p+1)i'} \\
        T + t_{at(p-1)i'} + t_{at(p+1)k'} + t_{i'k'} \\
        ~~~~~~~- t_{at(p-1)j'} - t_{at(p+1)j'} \leq T_{max}
    \end{cases}
\end{align}
the vertex $j'$ will be replaced by the edge $(i',k')$.

The combination of \eqref{eq:taylor}, \eqref{eq:replace11}, and \eqref{eq:replace21} can eliminate all the cases that the approximation of the objective after a replacement is worse than that value of the current route. However, we have to accept the trade-off between the running time and the quality of the \textit{Replace} operators in cases when the Taylor approximations are not close to their true values. Preliminary experiments show that this technique helps to reduce from 6.6\% to 79.1\% running time on average for small and medium datasets without decreasing the quality of the solution. For large-scale instances, when the time limit is reached, the ILS with Taylor approximations performs more iterations than the one without approximation technique; thus, it provides better objective values. 


\section{Experiments}\label{sec:Experiment}
\subsection{Experimental settings}
In the following, we conduct computational experiments to evaluate the performance of the exact methods and the metaheuristic proposed in the previous sections. A 64-bit Windows machine, with AMD Ryzen 7 3700X 8-Core Processor, 3.60 GHz, and 16GB of RAM, is used to run the experiments. Our algorithms are implemented in C++ and we link to IBM ILOG-CPLEX 12.10  to design the exact methods under default settings. It is worth noting that there was only 1 thread being used with a limited run time of 3600 seconds for each instance of the exact methods. In other words, the exact algorithms are forced to stop if the time budget is exceeded, and the best solutions are reported. The ILS terminates when the maximum number of iterations $nb_{iter} = 10,000$ is reached or when the time limit of 3600 seconds is attained, similarly to the exact methods for a fair comparison. On each instance, the heuristic method is run 20 times and the best solutions found and the average running time are recorded.

Since there is no benchmark instances available in the literature for the MCP-R, we generate a new set of benchmark instances to test the performance of the methods. We reuse the OP instances provided by \cite{ANGELELLI2014404, ANGELELLI2017} available at: \url{https://or-brescia.unibs.it/instances} from which we take the coordinates of locations, the distance matrix between any two vertices, and the value of $T_{max}$. In total, 32 instances of the OP with the number of vertices ranging from 14 to 299 have been selected to generate the MCP-R instances. To create customer utilities, we use instances from three datasets \texttt{ORlib, HM14}, and \texttt{NYC}, which are widely used in prior MCP studies. The set \texttt{ORlib} contains 15 instances with the number of zones varying in \{50, 100, 200, 400, 800\} and the number of locations in \{25, 50, 100\}, while \texttt{HM14} includes 3 instances with 1000 zones and 100 locations. The last set \texttt{NYC} has a single instance coming from a large-scale park-and-ride location problem in New York City, with 82,341 customer zones and 58 locations. This is the largest and most challenging instance in the context of the MCP. We also employ the same parameter settings as in previous studies \citep{Ljubic2018outer,MaiLodi2020_OA, dam2022submodularity}. The utility of location $i \in [m]$ to a zone $n\in \cN$ is computed as $V_{ni}=e^{v_{ni}-v_{n}}$ where $v_{ni}$ and $v_{n}$ are taken from the MCP instances and present deterministic parts of the random utilities related to each pair (zone, location) and competitors in zone, respectively.

The number of locations in the MCP data is modified to match the number of vertices in the OP data. For example, 14 locations and their corresponding utilities are randomly sampled from the \texttt{ORlib} instance of 25, 50, or 100 locations, and then they are combined with the coordinates of vertices in the OP instance \texttt{burma14} containing 14 vertices to create a new MCP-R instance. Note that the maximum number of locations in the MCP datasets is 100. Thus, for the OP instances with over 100 vertices, we merge two HM14 instances to form a new MCP instance. We then randomly choose utilities from that MCP instance until reaching to the required number of vertices as in the OP instance. In addition, for each OP instance, we create three R-MCP instances with different values of time budget: $T_{max}$ as in the original instance, $2T_{max}$, and $3T_{max}$. 

We compare our main algorithms, i.e., the nested cutting plane (\textbf{NCP}) and nested B\&C (denoted as \textbf{N-B\&C}) proposed in Section \ref{sec: CP and BC}, and the ILS developed in Section \ref{sec:Heuristic}, with the following baselines:
\begin{itemize}
    \item The \textbf{MILP} and \textbf{ Conic} reformulations presented in \eqref{prb:MILP}  and \eqref{prb:CONIC}. For both reformulations, we use $w^u = 1$ and $w_l = 0$ as valid upper and lower bounds for linearization, respectively. We note that in constraint \eqref{eq:linear1}, we use $w^{U}_{n} = 1\backslash (1+V_{ni})$ for each $i \in [m]$ due to the fact that $s_{ni}=w^{U}_{n}$ only if $x_{i}=1$. 
    \item Cutting Plane with MTZ constraints (\textbf{CP-MTZ}): a cutting plane algorithm with outer-approximation and submodular cuts, but instead of using sub-tour elimination cuts, we directly use MTZ constraints, i.e., Constraints \eqref{eq:start-end}--\eqref{eq:subtour}.
    \item B\&C with MTZ constraints (\textbf{B\&C-MTZ}): a B\&C algorithm with MTZ constraints, instead of using sub-tour elimination cuts.
    \item \textbf{ILS}: because of the randomness embedded in the heuristic, the ILS is executed 20 times on each instance to measure its stability. Our experiment on the ILS shows that the gap between the best-found objective and the average objective value is always less than $1.35\%$. This can prove that the heuristic is stable. Thus, we report the number of best objective values and the average runtime only.
\end{itemize}

The implementation of our cutting plane and  B\&C methods require partitioning the set of customer zones $\cN$ into disjoint groups, such that at each iteration, one cut per group is added to the master problem. The selection of the number of groups $\cL$  is crucial to strike a balance between the number of cuts that need to be added to the master problem at each iteration and the number of iterations required for the algorithms to converge. As shown in \cite{MaiLodi2020_OA}, increasing $\cL$ causes the master problem to grow faster in size, in terms of the number of constraints (resulting in longer solving times), but it would aid in reducing the number of iterations needed for convergence. In our context, we opt for $\cL = 20$ as it yields the best overall performance. Subsequently, we conduct experiments to assess the impact of $\cL$ on the cutting plane and B\&C's performance.


\subsection{Comparison Results}

\begin{table}[h!]
\centering
\resizebox{\textwidth}{!}{%
\begin{tabular}{c|ccc|ccc|ccc|cc}
          &       &        &               & \multicolumn{3}{c|}{MTZ}        & \multicolumn{3}{c|}{Nested}     &              &      \\
          & \multicolumn{3}{c|}{MILP/Conic} & \multicolumn{3}{c|}{CP/B\&C}   & \multicolumn{3}{c|}{NCP/N-B\&C} & \multicolumn{2}{c}{ILS} \\ \cline{1-12} 
Instances & \#Opt & \#Best & Time          & \#Opt & \#Best & Time          & \#Opt & \#Best & Time          & \#Opt/\#Best & Time \\
\hline
burma14   & \textbf{45}/37 & \textbf{45}/39  & 51.0/883.7    & \textbf{45/45} & \textbf{45/45}  & 1.4/0.8       & \textbf{45/45} & \textbf{45/45}  & 0.1/0.0       & \textbf{45/45}         & 4.9  \\
ulysses16 & 44/23 & \textbf{45}/31  & 413.3/2305.4  & 44/\textbf{45} & 44/\textbf{45}  & 101.5/8.5     & \textbf{45/45} & \textbf{45/45}  & 0.1/0.1       & \textbf{45/45}         & 6.0  \\
gr17      & \textbf{45}/27 & \textbf{45}/28  & 87.7/1831.7   & \textbf{45/45} & \textbf{45/45}  & 13.7/9.9      & \textbf{45/45} & \textbf{45/45}  & 0.1/0.1       & \textbf{45/45}         & 5.9  \\
gr21      & \textbf{45}/15 & \textbf{45}/30  & 191.7/2512.8  & \textbf{45/45} & \textbf{45/45}  & 24.9/35.0     & \textbf{45/45} & \textbf{45/45}  & 0.2/0.1       & \textbf{45/45}         & 7.0  \\
ulysses22 & 22/7  & 29/14  & 2450.8/3139.0 & 34/34 & 42/43  & 1300.2/1171.6 & \textbf{45/45} & \textbf{45/45}  & 0.1/0.1       & \textbf{45/45}         & 8.4  \\
gr24      & 43/6  & 44/16  & 461.\textbf{3/3}168.1  & 44/44 & 43/44  & 684.1/685.3   & \textbf{45/45} & \textbf{45/45}  & 0.3/0.1       & \textbf{45/45}         & 8.4  \\
fri26     & 27/0  & 29/10  & 741.4/-       & 9/10  & 26/15  & 2938.2/2467.5 & \textbf{30/30} & \textbf{30/30}  & 0.2/0.1       & \textbf{30/30}         & 8.3  \\
bays29    & 28/4  & 28/12  & 929.9/3042.5  & \textbf{30/30} & \textbf{30/30}  & 121.9/229.3   & \textbf{30/30} & \textbf{30/30}  & 0.5/0.2       & \textbf{30/30}         & 10.2 \\
dantzig42 & 0/0   & 1/1    & -/-           & 0/0   & 11/10  & -/-           & \textbf{30/30} & \textbf{30/30}  & 4.3/2.3       & \textbf{30/30}         & 14.9 \\
swiss42   & 14/2  & 16/5   & 2633.8/3345.3 & 29/\textbf{30} & \textbf{30/30}  & 485.8/418.2   & \textbf{30/30} & \textbf{30/30}  & 1.4/0.8       & \textbf{30/30}         & 16.9 \\
att48     & 0/0   & 0/0    & -/-           & 3/2   & 16/5   & 3431.2/3502.9 & \textbf{30/30} & \textbf{30/30}  & 3.5/2.5       & \textbf{30/30}         & 18.8 \\
gr48      & 12/0  & 12/1   & 2921.8/-      & 14/9  & 24/10  & 2638.6/2943.3 & \textbf{30/30} & \textbf{30/30}  & 4.9/3.6       & \textbf{30/30}         & 18.7 \\
hk48      & 1/0   & 3/0    & 3538.9/-      & 3/1   & 16/6   & 3418.9/3584.8 & \textbf{30/30} & \textbf{30/30}  & 6.9/3.1       & \textbf{30/30}         & 17.4 \\
eil51     & 8/0   & 8/0    & 2240.0/-      & 13/10 & 13/10  & 1249.8/2081.5 & \textbf{15/15} & \textbf{15/15}  & 7.1/4.9       & \textbf{15/15}         & 19.5 \\
berlin52  & 2/0   & 3/0    & 3429.8/-      & 14/6  & 14/8   & 1048.2/2592.6 & \textbf{15/15} & \textbf{15/15}  & 3.1/4.0       & \textbf{15/15}         & 22.5 \\
brazil58  & 0/0   & 0/0    & -/-           & 1/0   & 1/0    & 3453.1/-      & \textbf{15/15} & \textbf{15/15}  & 21.8/13.3     & \textbf{15/15}         & 26.3 \\
st70      & 0/0   & 0/0    & -/-           & 0/0   & 0/0    & -/-           & \textbf{15/15} & \textbf{15/15}  & 97.1/82.9     & \textbf{15/15}         & 33.0 \\
eil76     & 0/0   & 0/0    & -/-           & 2/0   & 3/0    & 2945.2/-      & \textbf{15/15} & \textbf{15/15}  & 217.\textbf{3/3}67.5   & \textbf{15/15}         & 36.8 \\
pr76      & 0/0   & 0/0    & -/-           & 0/0   & 0/0    & -/-           & \textbf{11/11} & 11/11  & 1246.3/1344.5 & 10/\textbf{14}         & 36.3 \\
gr96      & 0/0   & 0/0    & -/-           & 0/0   & 0/0    & -/-           & \textbf{14}/12 & \textbf{15}/12  & 1019.9/1364.5 & 9/10         & 57.9 \\
rat99     & 0/0   & 0/0    & -/-           & 0/0   & 0/0    & -/-           & \textbf{6/6}   & 7/7    & 2561.5/3251.7 & 4/\textbf{10}          & 52.6 \\ \hline
& & & & & & & \textbf{586}/584   & 588/584    & 247.5/307.0 & 578/\textbf{589}          & 20.2
\end{tabular}%
}
\caption{Comparison results for instances from the ORlib dataset}
\label{tab:table1}
\end{table}

Table \ref{tab:table1} presents comparison results for instances from the ORlib dataset. The first column lists groups of instances with the number of vertices varying from 14 to 99. The second, third, and fourth columns report the results of the \textbf{MILP} and \textbf{Conic} formulations. Here, Columns ``\#Opt'' and  ``\#Best" indicate the numbers of optimal and best solutions found by the \textbf{MILP} and \textbf{Conic} within the time limit, respectively. Column ``Time" refers to as the average running time (in seconds) for each group of instances. The symbol ``-'' means that the method cannot terminate within the time limit (3600 seconds). Similarly, the next blocks of columns display the numerical results for the MTZ-based formulations (\textbf{CP-MTZ} and \textbf{BC-MTZ}) and nested methods (\textbf{NCP} and \textbf{N-B\&C}). The last two columns present the results for the \textbf{ILS}, where Column ``\#Opt/\#Best'' reports the number of solutions that are proved to be optimal provided by the exact methods and the number of best solutions found by the \textbf{ILS}. The highest numbers of best and optimal solutions found over all methods are highlighted in bold. The last row presents the total numbers of optimal and best-found solutions (Columns \#Opt, \#Best, and \#Opt/\#Best) and average running time (Columns Time) over all instances for each method. Here, we will not report results for the methods that are clearly worse than the others.

As shown in Table \ref{tab:table1}, the \textbf{MILP} and \textbf{Conic} perform poorly, except when the number of locations is small, and the \textbf{MILP} outperforms the \textbf{Conic} in terms of both the number of optimal and best solutions. The \textbf{MILP} is also the faster method. The \textbf{CP-MTZ} and \textbf{BC-MTZ} perform better than the \textbf{MILP} and \textbf{Conic}, thanks to the utilization of outer-approximation and submodular cuts to approximate the nonlinear objective function. The \textbf{CP-MTZ} performs better than the \textbf{BC-MTZ}, especially on the larger instances. Two methods, \textbf{NCP} and \textbf{N-B\&C}, are superior compared to other exact methods, finding 586 and 584 optimal solutions out of 600 instances. The \textbf{NCP} and \textbf{N-B\&C} are even about 100 to 1000 times faster than other exact methods on some instances. The \textbf{ILS} algorithm can find 578 optimal solutions and provide the highest number of best solutions (589/600 instances). Its computation time is also quite fast, less than 1 minute. 

\begin{table}[]
\centering
\resizebox{\textwidth}{!}{%
\begin{tabular}{c|ccc|ccc|ccc|cc}
 & \multicolumn{3}{c|}{} & \multicolumn{3}{c|}{MTZ} & \multicolumn{3}{c|}{Nested} & \multicolumn{2}{c}{} \\
 & \multicolumn{3}{c|}{MILP/Conic} & \multicolumn{3}{c|}{CP/B\&C} & \multicolumn{3}{c|}{NCP/N-B\&C} & \multicolumn{2}{c}{ILS} \\ \cline{1-12} 
Instances & \#Opt & \#Best & Time & \#Opt & \#Best & Time & \#Opt & \#Best & Time & \#Opt/\#Best & Time \\ \hline
burma14 & \textbf{9}/8 & \textbf{9}/8 & 226.0/1795.2 & \textbf{9/9} & \textbf{9/9} & 1.0/1.4 & \textbf{9/9} & \textbf{9/9} & 0.1/0.9 & \textbf{9/9} & 12.8 \\
ulysses16 & 6/2 & \textbf{9}/3 & 1439.9/2856.5 & \textbf{9/9} & \textbf{9/9} & 19.8/23.2 & \textbf{9/9} & \textbf{9/9} & 0.1/3.8 & \textbf{9/9} & 15.0 \\
gr17 & \textbf{9}/0 & \textbf{9}/0 & 404.0/- & \textbf{9/9} & \textbf{9/9} & 5.7/37.2 & \textbf{9/9} & \textbf{9/9} & 0.1/1.6 & \textbf{9/9} & 15.0 \\
gr21 & 4/0 & 5/0 & 2482.5/- & \textbf{9/9} & \textbf{9/9} & 22.5/60.4 & \textbf{9/9} & \textbf{9/9} & 0.2/13.1 & \textbf{9/9} & 18.1 \\
ulysses22 & 2/0 & 2/0 & 3115.3/- & 5/6 & \textbf{9}/8 & 2335.3/1506.9 & \textbf{9/9} & \textbf{9/9} & 0.2/16.7 & \textbf{9/9} & 21.4 \\
gr24 & 2/0 & 2/0 & 3404.5/- & \textbf{9}/5 & \textbf{9}/8 & 674.0/1782.1 & \textbf{9/9} & \textbf{9/9} & 0.7/19.3 & \textbf{9/9} & 19.8 \\
fri26 & 3/0 & 3/0 & 2868.1/- & 2/3 & 8/4 & 3240.7/2944.7 & \textbf{9/9} & \textbf{9/9} & 0.5/16.9 & \textbf{9/9} & 19.9 \\
bays29 & 0/0 & 0/0 & \textbf{-/-} & \textbf{9/9} & \textbf{9/9} & 68.2/694.2 & \textbf{9/9} & \textbf{9/9} & 0.6/15.3 & \textbf{9/9} & 24.2 \\
dantzig42 & 0/0 & 0/0 & \textbf{-/-} & 0/0 & 7/1 & \textbf{-/-} & \textbf{9/9} & \textbf{9/9} & 5.7/8.3 & \textbf{9/9} & 38.0 \\
swiss42 & 0/0 & 0/0 & \textbf{-/-} & \textbf{9}/6 & \textbf{9/9} & 314.6/2170.3 & \textbf{9/9} & \textbf{9/9} & 3.1/3.8 & 8/8 & 42.6 \\
att48 & 0/0 & 0/0 & \textbf{-/-} & 0/0 & 4/1 & \textbf{-/-} & \textbf{9/9} & \textbf{9/9} & 7.4/9.1 & 8/8 & 48.2 \\
gr48 & 0/0 & 0/0 & \textbf{-/-} & 4/0 & \textbf{7/3} & 2248.1/- & \textbf{9/9} & \textbf{9/9} & 11.3/22.7 & \textbf{9/9} & 48.4 \\
hk48 & 0/0 & 0/0 & \textbf{-/-} & 1/0 & 4/0 & 3458.3/- & \textbf{9/9} & \textbf{9/9} & 2.7/15.7 & 5/5 & 42.3 \\
eil51 & 0/0 & 0/0 & \textbf{-/-} & \textbf{9}/5 & \textbf{9}/7 & 592.3/- & \textbf{9/9} & \textbf{9/9} & 4.6/9.7 & 6/6 & 48.9 \\
berlin52 & 0/0 & 0/0 & \textbf{-/-} & 8/2 & \textbf{9}/5 & 775.6/2972.1 & \textbf{9/9} & \textbf{9/9} & 4.1/8.2 & \textbf{9/9} & 58.0 \\
brazil58 & 0/0 & 0/0 & \textbf{-/-} & 3/0 & 3/0 & 2315.7/- & \textbf{9/9} & \textbf{9/9} & 8.4/78.6 & 8/8 & 72.2 \\
st70 & 0/0 & 0/0 & \textbf{-/-} & 0/0 & 1/0 & \textbf{-/-} & \textbf{9/9} & \textbf{9/9} & 206.4/273.4 & \textbf{9/9} & 81.1 \\
eil76 & 0/0 & 0/0 & \textbf{-/-} & 0/0 & 6/0 & 3293.8/- & \textbf{9/9} & \textbf{9/9} & 41.3/57.5 & 2/2 & 95.4 \\
pr76 & 0/0 & 0/0 & \textbf{-/-} & 0/0 & 0/0 & \textbf{-/-} & \textbf{6}/4 & \textbf{6}/4 & 622.0/2621.9 & 2/5 & 88.7 \\
gr96 & 0/0 & 0/0 & \textbf{-/-} & 0/0 & 0/0 & \textbf{-/-} & \textbf{9}/8 & \textbf{9}/8 & 219.2/1411.0 & 5/5 & 146.9 \\
rat99 & 0/0 & 0/0 & \textbf{-/-} & 0/0 & 0/0 & \textbf{-/-} & \textbf{6}/2 & \textbf{7}/2 & 1573.2/3111.9 & 3/4 & 128.3 \\
\hline
 &  &  &  &  &  &  & \textbf{183}/176 & \textbf{184}/176 & 129.1/367.6 & 155/159 & 51.7
\end{tabular}%
}
\caption{Comparison results for instances from the HM14 dataset}
\label{tab:table2}
\end{table}

We switch to the instances generated from the \texttt{HM14} dataset. In Table \ref{tab:table2}, we report numerical results for these instances. The obtained results show the same phenomenon as in the last dataset in terms of the relative performance of the methods. The \textbf{NCP} and \textbf{N-B\&C} remain the two strongest methods. The \textbf{NCP} is slightly better than the \textbf{N-B\&C} with 183/189 optimal solutions found, compared to 176/189 optimal solutions given by the \textbf{N-B\&C}. The \textbf{ILS} is still the third approach regarding to the capability of providing the best solutions, followed by \textbf{CP-MTZ}, \textbf{B\&C-MTZ}, \textbf{MILP}, and \textbf{Conic} approaches, respectively. In addition, as the number of zones increases, we observe that the runtime of all methods significantly increases, compared to those reported in Table \ref{tab:table1}. In particular, the computation time of \textbf{ILS}  increases the most due to the more complexity in calculating the objective function. Our experiment also shows that the nested approaches work less effectively on the instances with smaller time budgets $T_{max}$. However, the performance of the ILS decreases when $T_{max}$ increases. A possible reason is that, when the time budget increases, more locations could be selected and the OP tends to become the TSP which is easier to solve for the \textbf{NCP} and \textbf{N-B\&C}. In contradiction, augmenting $T_{max}$ leads to more feasible insertions that the \textbf{ILS} has to consider. As a consequence, the \textbf{ILS} needs more time to process its insertion operation, reducing the duration for exploring new solution spaces. 

\begin{table}[!h]
\centering
\resizebox{0.7\textwidth}{!}{%
\begin{tabular}{c|ccc|ccc|cc}
          & \multicolumn{3}{c|}{MTZ}        & \multicolumn{3}{c|}{Nested}     &               &         \\
          & \multicolumn{3}{c|}{CP/B\&C}    & \multicolumn{3}{c|}{NCP/N-B\&C} & \multicolumn{2}{c}{ILS} \\ \hline
Instances & \#Opt & \#Best & Time          & \#Opt  & \#Best & Time         & \#Opt/\#Best  & Time    \\ \hline
burma14   & \textbf{\textbf{3/3}}   & \textbf{\textbf{3/3}}    & 12.2/2.6      & \textbf{\textbf{3/3}}    & \textbf{3/3}    & 0.5/2.0      & \textbf{\textbf{3/3}}           & 3145.3  \\
ulysses16 & \textbf{\textbf{3/3}}   & \textbf{\textbf{3/3}}    & 6.2/4.5       & \textbf{\textbf{3/3}}    & \textbf{\textbf{3/3}}    & 0.7/3.3      & \textbf{\textbf{3/3}}           & 3466.5  \\
gr17      & \textbf{\textbf{3/3}}   & \textbf{\textbf{3/3}}    & 8.8/15.5      & \textbf{\textbf{3/3}}    & \textbf{\textbf{3/3}}    & 0.6/2.9      & \textbf{\textbf{3/3}}           & 3587.8  \\
gr21      & \textbf{\textbf{3/3}}   & \textbf{\textbf{3/3}}    & 9.8/12.2      & \textbf{\textbf{3/3}}    & \textbf{\textbf{3/3}}    & 1.1/4.0      & \textbf{\textbf{3/3}}           & -       \\
ulysses22 & \textbf{\textbf{3/3}}   & \textbf{3/3}    & 234.1/118.3   & \textbf{3/3}    & \textbf{3/3}    & 0.9/9.4      & \textbf{3/3}           & -       \\
gr24      & 2/\textbf{3}   & \textbf{3/3}    & 1518.2/937.7  & \textbf{3/3}    & \textbf{3/3}    & 1.4/6.1      & \textbf{3/3}           & -       \\
fri26     & 1/1   & \textbf{3/3}    & 2478.5/2460.7 & \textbf{3/3}    & \textbf{3/3}    & 1.5/5.4      & \textbf{3/3}           & 3439.1  \\
bays29    & \textbf{3/3}   & \textbf{3/3}    & 713.4/797.1   & \textbf{3/3}    & \textbf{3/3}    & 2.3/9.2      & \textbf{3/3}           & 3565.9  \\
dantzig42 & 1/1   & \textbf{3/3}    & 3309.8/3293.3 & \textbf{3/3}    & \textbf{3/3}    & 34.0/39.0    & \textbf{3/3}           & -       \\
swiss42   & \textbf{3/3}   & \textbf{3/3}    & 440.6/1201.6  & \textbf{3/3}    & \textbf{3/3}    & 2.6/32.9     & \textbf{3/3}           & -       \\
att48     & 0/0   & 0/1    & -/-           & \textbf{3/3}    & \textbf{3/3}    & 12.6/55.2    & \textbf{3/3}           & -       \\
gr48      & 0/0   & \textbf{3}/1    & -/-           & \textbf{3/3}    & \textbf{3/3}    & 7.4/71.5     & \textbf{3/3}           & -       \\
hk48      & 1/1   & \textbf{3/3}    & 2873.0/2985.4 & \textbf{3/3}    & \textbf{3/3}    & 21.5/73.1    & \textbf{3/3}           & -       \\
eil51     & 1/0   & \textbf{3}/2    & 2914.5/-      & \textbf{3/3}    & \textbf{3/3}    & 8.6/74.0     & \textbf{3/3}           & -       \\
berlin52  & 1/0   & \textbf{3}/2    & 2382.7/-      & \textbf{3/3}    & \textbf{3/3}    & 12.1/67.2    & \textbf{3/3}           & -       \\
brazil58  & 0/0   & 1/1    & -/-           & 2/\textbf{3}    & 2/\textbf{3}    & 1421.6/440.0 & \textbf{3/3}           & -       \\ \hline
&    &     &            & 47/\textbf{48}    & 47/\textbf{48}    & 95.6/56.0 & \textbf{48}/\textbf{48}           & 3556.5            
\end{tabular}%
}
\caption{Comparison results for instances from the NYC dataset}
\label{tab:table3}
\end{table}

Table \ref{tab:table3} presents the comparison results on the most challenging dataset, \texttt{NYC}. This dataset contains instances with more than 80,000 zones and up to 59 locations. Preliminary experiments show that the \textbf{MILP} and \textbf{Conic} approaches are too slow for these demanding instances, therefore, only five remaining approaches are examined. Although the \textbf{ILS}, along with \textbf{N-B\&C}, seems to perform the best in terms of the number of optimal solutions found, \textbf{ILS} is worse regarding the runtime. The complexity of calculating the objective function in \textbf{ILS} dramatically increases when the number of zones becomes large, obviously slowing down the algorithm significantly. Both nested-based approaches solve the problem quickly in almost all instances. The \textbf{NCP} performs slightly worse than the \textbf{N-B\&C}. Between the \textbf{CP-MTZ} and \textbf{B\&C-MTZ}, the performance of the former is better than that of the latter. These two methods are able to find the best objective values for instances of up to 42 locations. 

To further assess the performance of our methods, we test them on larger instances with 1000 customer zones and more than 100 locations. Table \ref{tab:table4} below reports the obtained results. Clearly, we observe that the \textbf{ILS} achieves the best performance with respect to the number of best-found solutions and computational time. Over 24 instances, it provides 19 best-found solutions including 10 optimal solutions, while the second best method, \textbf{NCP}, only finds 14 best-found solutions in which 11 solutions are optimal. Moreover, the \textbf{ILS} requires less runtime than other methods to solve all the considering instances, except for the largest one \texttt{pr299}. 

\begin{table}[!h]
\centering
\resizebox{0.7\textwidth}{!}{%
\begin{tabular}{c|ccc|ccc|cc}
         & \multicolumn{3}{c|}{MTZ}        & \multicolumn{3}{c|}{Nested}     &              &      \\
          & \multicolumn{3}{c|}{CP/B\&C}   & \multicolumn{3}{c|}{NCP/N-B\&C} & \multicolumn{2}{c}{ILS} \\ \hline 
Instances & \#Opt & \#Best & Time          & \#Opt & \#Best & Time          & \#Opt/\#Best & Time \\ \hline
bier127 & 1/0   & 1/0    & 2530.68/- & \textbf{3}/2   & \textbf{3}/\textbf{3}    & 1456.13/2142.68 & 2/2          & 1355.64 \\
pr152   & 0/0   & 0/0    & -/-       & \textbf{2}/1   & \textbf{3}/1    & 1857.35/2761.18 & \textbf{2}/2          & 389.80  \\
brg180  & 0/0   & 0/0    & -/-       & \textbf{2}/2   & 2/2    & 1212.39/1319.41 & \textbf{2}/\textbf{3}          & 633.39  \\
rat195  & 0/0   & 0/0    & -/-       & \textbf{1}/0   & 1/1    & 2422.29/-       & \textbf{1}/\textbf{3}          & 2054.30 \\
gr229   & 0/0   & 0/0    & -/-       & \textbf{1}/\textbf{1}   & \textbf{2}/1    & 2419.64/3530.53 & \textbf{1}/\textbf{2}          & 1463.81 \\
pr264   & 0/0   & 0/0    & -/-       & \textbf{2}/0   & \textbf{2}/0    & 2714.29/-       & 1/\textbf{2}          & 1736.82 \\
a280    & 0/0   & 0/0    & -/-       & 0/0   & 0/0    & -/-             & 0/\textbf{3}          & 2472.29 \\
pr299   & 0/0   & 0/0    & -/-       & \textbf{1}/0   & 1/1    & 2898.01/-       & \textbf{1}/\textbf{2}          & 3584.52\\ \hline
   &    &     &     & \textbf{11}/6   & 14/9    &  2323.04/3019.22      & 10/\textbf{19}          & 1711.32\\
\end{tabular}%
}
\caption{Comparison results for large  instances with utilities from HM14 dataset}
\label{tab:table4}
\end{table}

\subsection{Impact of $\cL$ on the Performance of the NCP}

\begin{figure}[htb]
  \includegraphics[width=0.9\textwidth]{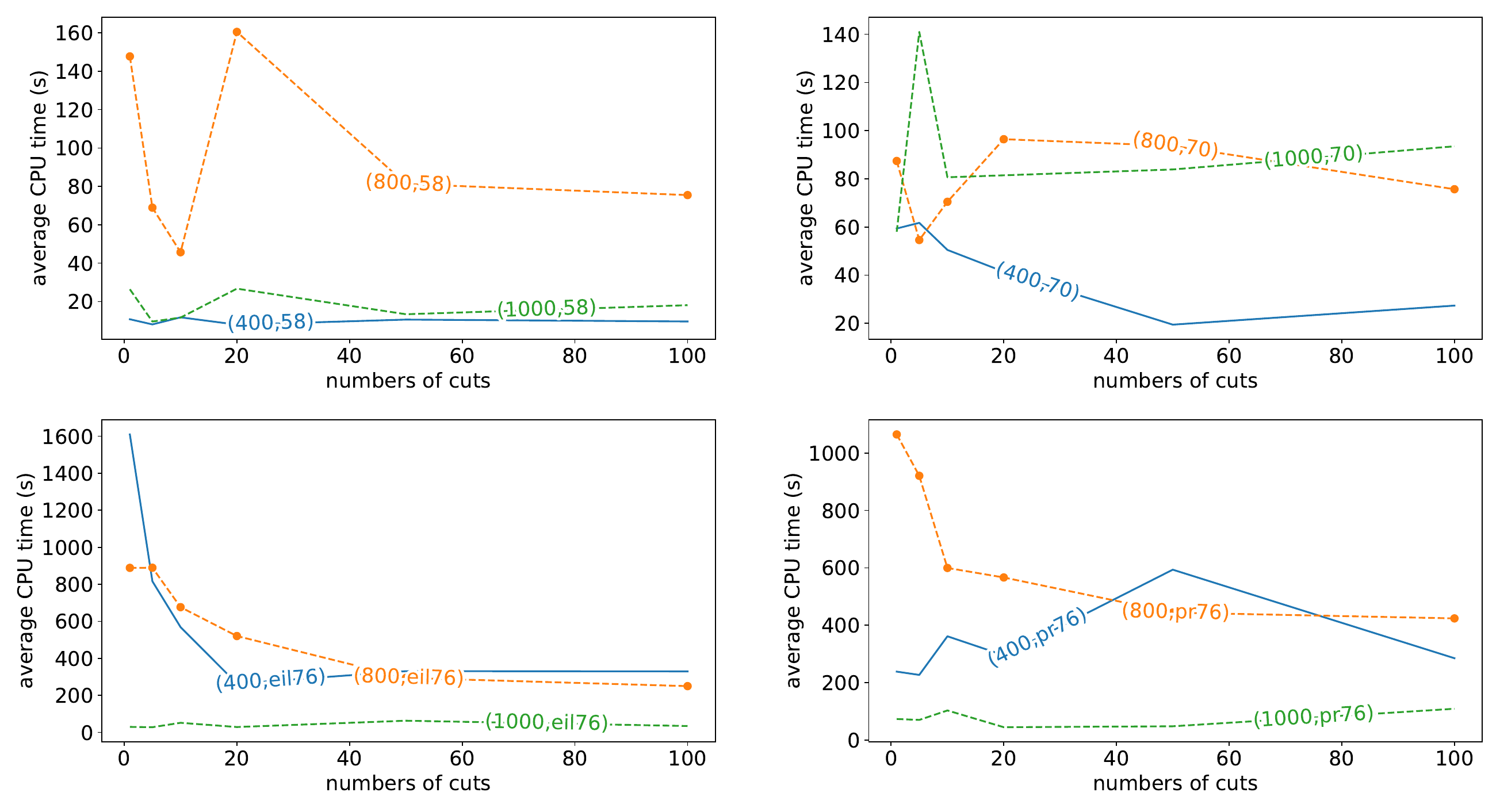}
  \caption{{Average computational time  of  \textbf{NCP} as a function of $\cL$.}}
  \label{fig:CP-cut}
\end{figure}

\begin{figure}[htb]
  \includegraphics[width=0.9\textwidth]{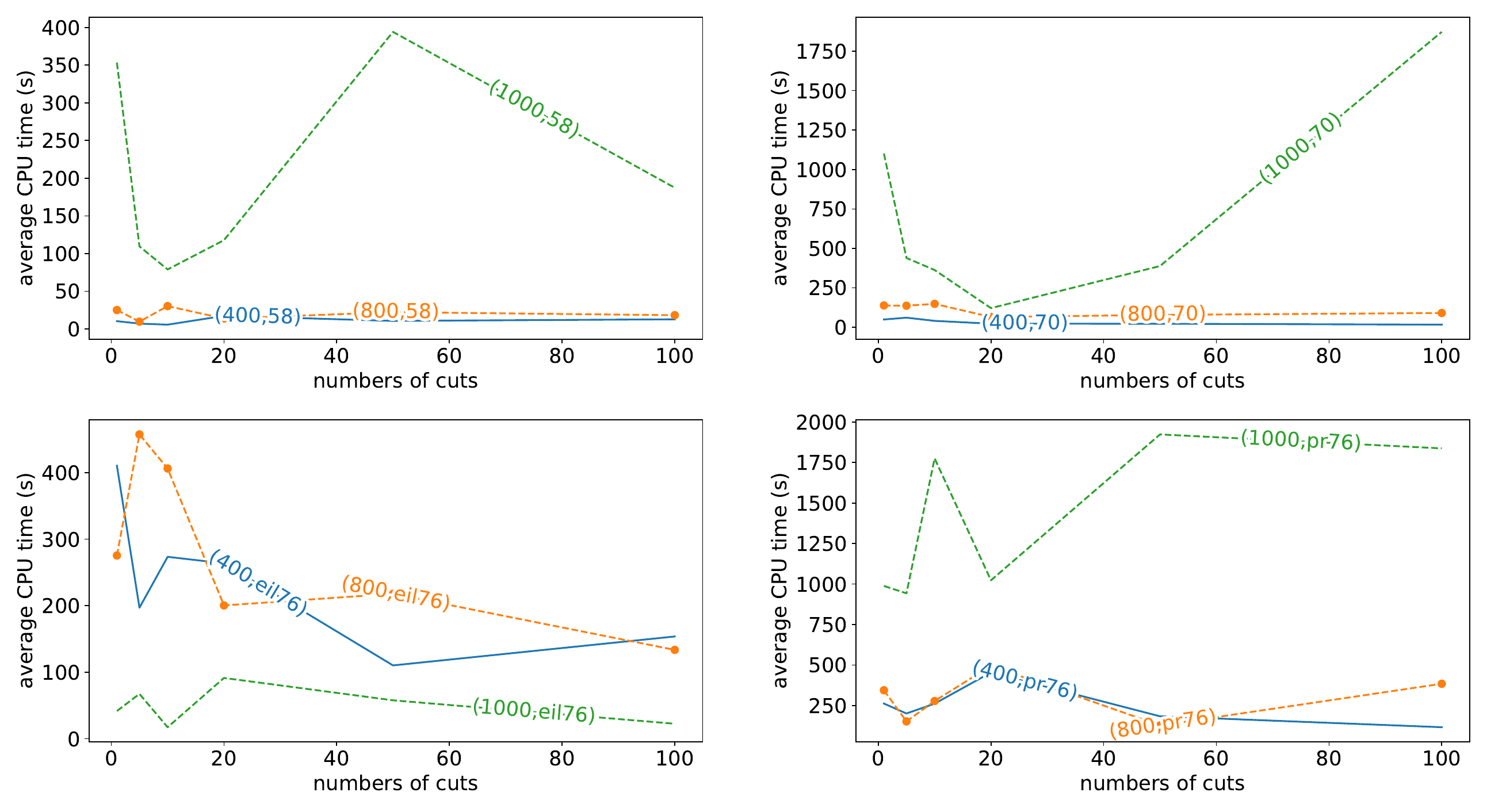}
  \caption{{Average computational time  of  \textbf{N-B\&C} as a function of $\cL$. }}
  \label{fig:BnC-cut}
\end{figure}

In this section, we examine the performance of our cutting plane and B\&C algorithms with varying number of groups $\cL$ (or the number of outer-approximation and submodular cuts added to the master problem at each iteration). To this end, we select four representative OP instances \texttt{brazil58, st70, eil76, pr76}, where the number of available locations are 58, 70, 76, 76, respectively. More precisely, for each OP instance, we generate four MCP-R instances by varying the number of customer zones in the set $\{400,800,1000\}$. As such, 12 instances have been used for the experiment. For each instance, we test the \textbf{NCP} and \textbf{N-B\&C} with different values of $\cL$  taken from the set $\{1,5,10,20,50,100\}$. 

In Figure \ref{fig:CP-cut}, we report the average computation times of the \textbf{CP-B\&C} method when $\cL$ varies, noting that all the instances are solved to optimality. The figure shows that  the runtime exhibits instability for $\cL\leq 50$, but  execution time tends to be decreasing and stable  when $\cL\geq 50$. Specifically, when the number of cuts increases from 50 to 100, the runtime tends to experience a slight upward trend  and the \textbf{NCP} method achieves the best performance with $\cL = 50$. On the other hand, Figure \ref{fig:BnC-cut} shows the computational times of the \textbf{N-B\&C} method when $\cL$ varies. The computation times are unstable but tend to decrease rapidly as $\cL$ increases from 1 to 20, and then increases again as it goes from 20 to 100. We observe that the \textbf{N-B\&C} method achieves the best performance with $\cL = 20$.

\section{Conclusion}\label{sec:concl}

In this paper, we have considered a competitive facility location problem with routing constraints, where customer behavior is modeled using a MNL discrete choice model. The problem under consideration holds practical relevance in real-life applications and is characterized by the nonlinearity of the objective function and complex routing constraints, making existing approaches inefficient. To address this challenging problem, we have explored three types of cuts --  outer-approximation, submodular, and sub-tour elimination to develop new cutting plane and branch-and-cut approaches, which involve adding cuts to a master problem through two nested loops. We theoretically demonstrated that our nested cutting plane method consistently converges to optimal solutions after a finite number of iterations. Additionally, we have developed a metaheuristic based on the ILS to enhance the capability of solving large instances. However, the limit of the metaheuristic appears when the number of customer zones increases. Experiments conducted on instances of various sizes demonstrate the superiority of our nested cutting plane and B\&C approaches, which can solve to optimality the instances with a very large number of customer zones. For the instances with numerous locations, the ILS-based metaheuristic is more recommended. However, the limit of the metaheuristic appears as the number of customer zones increases. Future works will be dedicated to exploring more advanced choice models, such as nested models or cross-nested models \citep{Trai03}, or incorporating more complex and practical routing constraints.

\bibliographystyle{informs2014}
\bibliography{reference}

\begin{thebibliography}{52}
\providecommand{\natexlab}[1]{#1}
\providecommand{\url}[1]{\texttt{#1}}
\providecommand{\urlprefix}{URL }

\bibitem[{Angelelli et~al.(2017)Angelelli, Archetti, Filippi, \protect\BIBand{} Vindigni}]{ANGELELLI2017}
Angelelli E, Archetti C, Filippi C, Vindigni M (2017) The probabilistic orienteering problem. \emph{Computers \& Operations Research} 81:269--281.

\bibitem[{Angelelli et~al.(2014)Angelelli, Archetti, \protect\BIBand{} Vindigni}]{ANGELELLI2014404}
Angelelli E, Archetti C, Vindigni M (2014) The clustered orienteering problem. \emph{European Journal of Operational Research} 238(2):404--414.

\bibitem[{Archetti et~al.(2018)Archetti, Carrabs, \protect\BIBand{} Cerulli}]{ARCHETTI2018}
Archetti C, Carrabs F, Cerulli R (2018) The set orienteering problem. \emph{European Journal of Operational Research} 267(1):264--272.

\bibitem[{Ben-Akiva(1973)}]{BenA73}
Ben-Akiva M (1973) \emph{The structure of travel demand models}. Ph.D. thesis, MIT.

\bibitem[{Ben-Akiva \protect\BIBand{} Bierlaire(1999)}]{BenABier99a}
Ben-Akiva M, Bierlaire M (1999) \emph{Discrete Choice Methods and their Applications to Short Term Travel Decisions}, 5--33 (Boston, MA: Springer US).

\bibitem[{Ben-Akiva \protect\BIBand{} Lerman(1985)}]{BenALerm85}
Ben-Akiva M, Lerman SR (1985) \emph{Discrete Choice Analysis: Theory and Application to Travel Demand} (MIT Press, Cambridge, Massachusetts).

\bibitem[{Benati \protect\BIBand{} Hansen(2002)}]{FLO_Benati2002maximum}
Benati S, Hansen P (2002) The maximum capture problem with random utilities: Problem formulation and algorithms. \emph{European Journal of Operational Research} 143(3):518--530.

\bibitem[{Bonami et~al.(2008)Bonami, Biegler, Conn, Cornu{\'e}jols, Grossmann, Laird, Lee, Lodi, Margot, Sawaya et~al.}]{OA_Bonami2008algorithmic}
Bonami P, Biegler LT, Conn AR, Cornu{\'e}jols G, Grossmann IE, Laird CD, Lee J, Lodi A, Margot F, Sawaya N, et~al. (2008) An algorithmic framework for convex mixed integer nonlinear programs. \emph{Discrete Optimization} 5(2):186--204.

\bibitem[{Calinescu et~al.(2007)Calinescu, Chekuri, P{\'a}l, \protect\BIBand{} Vondr{\'a}k}]{calinescu2007maximizing}
Calinescu G, Chekuri C, P{\'a}l M, Vondr{\'a}k J (2007) Maximizing a submodular set function subject to a matroid constraint. \emph{International Conference on Integer Programming and Combinatorial Optimization}, 182--196 (Springer).

\bibitem[{Croes(1958)}]{2opt}
Croes GA (1958) A method for solving traveling-salesman problems. \emph{Operations Research} 6(6):791--812.

\bibitem[{\c{S}en et~al.(2018)\c{S}en, Atamt\"{u}rk, \protect\BIBand{} Kaminsky}]{Sen2017}
\c{S}en A, Atamt\"{u}rk A, Kaminsky P (2018) A conic integer programming approach to constrained assortment optimization under the mixed multinomial logit model. \emph{Operations Research} 66(4):994--1003.

\bibitem[{Dam et~al.(2022)Dam, Ta, \protect\BIBand{} Mai}]{dam2022submodularity}
Dam TT, Ta TA, Mai T (2022) Submodularity and local search approaches for maximum capture problems under generalized extreme value models. \emph{European Journal of Operational Research} 300(3):953--965.

\bibitem[{Dam et~al.(2023)Dam, Ta, \protect\BIBand{} Mai}]{dam2023robust}
Dam TT, Ta TA, Mai T (2023) Robust maximum capture facility location under random utility maximization models. \emph{European Journal of Operational Research} 310(3):1128--1150.

\bibitem[{Dolinskaya et~al.(2018)Dolinskaya, Shi, \protect\BIBand{} Smilowitz}]{DOLINSKAYA2018}
Dolinskaya I, Shi ZE, Smilowitz K (2018) Adaptive orienteering problem with stochastic travel times. \emph{Transportation Research Part E: Logistics and Transportation Review} 109:1--19.

\bibitem[{Duran \protect\BIBand{} Grossmann(1986)}]{Duran1986}
Duran MA, Grossmann IE (1986) An outer-approximation algorithm for a class of mixed-integer nonlinear programs. \emph{Mathematical Programming} 36:307--339.

\bibitem[{Faigl \protect\BIBand{} Pěnička(2017)}]{Faigl2017}
Faigl J, Pěnička R (2017) On close enough orienteering problem with dubins vehicle. \emph{IEEE/RSJ International Conference on Intelligent Robots and Systems (IROS)}, 5646--5652.

\bibitem[{Feillet et~al.(2005)Feillet, Dejax, \protect\BIBand{} Gendreau}]{Feillet2005}
Feillet D, Dejax P, Gendreau M (2005) Traveling salesman problems with profits. \emph{Transportation Science} 39(2):188--205.

\bibitem[{Fischetti et~al.(1998)Fischetti, Gonz\'{a}lez, \protect\BIBand{} Toth}]{Fischetti}
Fischetti M, Gonz\'{a}lez JJS, Toth P (1998) Solving the orienteering problem through branch-and-cut. \emph{INFORMS Journal on Computing} 10(2):133--148.

\bibitem[{Fosgerau \protect\BIBand{} Bierlaire(2009)}]{FosgBier09}
Fosgerau M, Bierlaire M (2009) Discrete choice models with multiplicative error terms. \emph{Transportation Research Part B} 43(5):494--505.

\bibitem[{Freeman et~al.(2018)Freeman, Keskin, \protect\BIBand{} İbrahim Çapar}]{FREEMAN2018}
Freeman NK, Keskin BB, İbrahim Çapar (2018) Attractive orienteering problem with proximity and timing interactions. \emph{European Journal of Operational Research} 266(1):354--370.

\bibitem[{Freire et~al.(2016)Freire, Moreno, \protect\BIBand{} Yushimito}]{Freire2015}
Freire A, Moreno E, Yushimito W (2016) A branch-and-bound algorithm for the maximum capture problem with random utilities. \emph{European Journal of Operational Research} 252(1):204--212.

\bibitem[{Golden et~al.(1987)Golden, Levy, \protect\BIBand{} Vohra}]{Golden1987}
Golden BL, Levy L, Vohra R (1987) The orienteering problem. \emph{Naval Research Logistics (NRL)} 34(3):307--318.

\bibitem[{Gunawan et~al.(2015)Gunawan, Lau, \protect\BIBand{} Lu}]{Gunawan2015}
Gunawan A, Lau HC, Lu K (2015) An iterated local search algorithm for solving the orienteering problem with time windows. \emph{Evolutionary Computation in Combinatorial Optimization}, 61--73 (Cham: Springer International Publishing).

\bibitem[{Gunawan et~al.(2016)Gunawan, Lau, \protect\BIBand{} Vansteenwegen}]{GUNAWAN2016}
Gunawan A, Lau HC, Vansteenwegen P (2016) Orienteering problem: A survey of recent variants, solution approaches and applications. \emph{European Journal of Operational Research} 255(2):315--332.

\bibitem[{Haase(2009)}]{Haase2009}
Haase K (2009) Discrete location planning, \urlprefix\url{http://hdl.handle.net/2123/19420}.

\bibitem[{Haase \protect\BIBand{} M{\"u}ller(2014)}]{FLO_Haase2014comparison}
Haase K, M{\"u}ller S (2014) A comparison of linear reformulations for multinomial logit choice probabilities in facility location models. \emph{European Journal of Operational Research} 232(3):689--691.

\bibitem[{Jorgensen et~al.(2017)Jorgensen, Chen, Milam, \protect\BIBand{} Pavone}]{Jorgensen2017}
Jorgensen S, Chen RH, Milam MB, Pavone M (2017) The matroid team surviving orienteers problem: Constrained routing of heterogeneous teams with risky traversal. \emph{IEEE/RSJ International Conference on Intelligent Robots and Systems (IROS)}, 5622--5629.

\bibitem[{Jorgensen et~al.(2018)Jorgensen, Chen, Milam, \protect\BIBand{} Pavone}]{Jorgensen2018}
Jorgensen S, Chen RH, Milam MB, Pavone M (2018) The team surviving orienteers problem: Routing teams of robots in uncertain environments with survival constraints. \emph{Auton. Robots} 42(4):927–952.

\bibitem[{Lamontagne et~al.(2023)Lamontagne, Carvalho, Frejinger, Gendron, Anjos, \protect\BIBand{} Atallah}]{Lamontagne2023}
Lamontagne S, Carvalho M, Frejinger E, Gendron B, Anjos MF, Atallah R (2023) Optimising electric vehicle charging station placement using advanced discrete choice models. \emph{INFORMS Journal on Computing} 35(5):1195--1213.

\bibitem[{Laporte \protect\BIBand{} Martín(2007)}]{Laporte2007}
Laporte G, Martín IR (2007) Locating a cycle in a transportation or a telecommunications network. \emph{Networks} 50(1):92--108.

\bibitem[{Li(1994)}]{LI1994}
Li HL (1994) A global approach for general 0–1 fractional programming. \emph{European Journal of Operational Research} 73(3):590--596.

\bibitem[{Ljubi{\'c} \protect\BIBand{} Moreno(2018)}]{Ljubic2018outer}
Ljubi{\'c} I, Moreno E (2018) Outer approximation and submodular cuts for maximum capture facility location problems with random utilities. \emph{European Journal of Operational Research} 266(1):46--56.

\bibitem[{Mai \protect\BIBand{} Lodi(2020)}]{MaiLodi2020_OA}
Mai T, Lodi A (2020) A multicut outer-approximation approach for competitive facility location under random utilities. \emph{European Journal of Operational Research} 284(3):874--881.

\bibitem[{McFadden(1978)}]{McFa78}
McFadden D (1978) Modeling the choice of residential location. \emph{Spatial Interaction Theory and Residential Location}, 75--96 (Amsterdam: North-Holland).

\bibitem[{McFadden(2001)}]{Mcfadden2001economicNobel}
McFadden D (2001) Economic choices. \emph{American Economic Review} 351--378.

\bibitem[{McFadden \protect\BIBand{} Train(2000)}]{McFaTrai00}
McFadden D, Train K (2000) Mixed {MNL} models for discrete response. \emph{Journal of applied Econometrics} 447--470.

\bibitem[{Mehmanchi et~al.(2020)Mehmanchi, Gillen, G{\'o}mez, \protect\BIBand{} Prokopyev}]{mehmanchi2020robust}
Mehmanchi E, Gillen CP, G{\'o}mez A, Prokopyev OA (2020) On robust fractional 0-1 programming. \emph{INFORMS Journal on Optimization} 2(2):96--133.

\bibitem[{Mehmanchi et~al.(2019)Mehmanchi, Gómez, \protect\BIBand{} Prokopyev}]{Mehmanchi2019}
Mehmanchi E, Gómez A, Prokopyev O (2019) Fractional 0–1 programs: links between mixed-integer linear and conic quadratic formulations. \emph{Journal of Global Optimization} 75:273–339.

\bibitem[{Meyer \protect\BIBand{} Glock(2022)}]{Meyer2022}
Meyer F, Glock K (2022) Kinematic orienteering problem with time-optimal trajectories for multirotor uavs. \emph{IEEE Robotics and Automation Letters} 7(4):11402--11409.

\bibitem[{Miller et~al.(1960)Miller, Tucker, \protect\BIBand{} Zemlin}]{MTZ1960}
Miller CE, Tucker AW, Zemlin RA (1960) Integer programming formulation of traveling salesman problems. \emph{J. ACM} 7(4):326–329.

\bibitem[{Nemhauser \protect\BIBand{} Wolsey(1981)}]{nemhauser1981maximizing}
Nemhauser GL, Wolsey LA (1981) Maximizing submodular set functions: formulations and analysis of algorithms. \emph{North-Holland Mathematics Studies}, volume~59, 279--301 (Elsevier).

\bibitem[{Nguyen et~al.(2022)Nguyen, Luong, Hà, \protect\BIBand{} Ha-Bang}]{Nguyen2021AnEB}
Nguyen MA, Luong H, Hà M, Ha-Bang B (2022) An efficient branch-and-cut algorithm for the parallel drone scheduling traveling salesman problem. \emph{4OR} 21:609–637.

\bibitem[{Or(1976)}]{OrOpt}
Or I (1976) \emph{Traveling Salesman-Type Combinatorial Problems and their Relation to the Logistics of Regional Blood Banking}. Ph.D. thesis, Evanston, IL.

\bibitem[{Pěnička et~al.(2019)Pěnička, Faigl, \protect\BIBand{} Saska}]{Penicka2019}
Pěnička R, Faigl J, Saska M (2019) Physical orienteering problem for unmanned aerial vehicle data collection planning in environments with obstacles. \emph{IEEE Robotics and Automation Letters} 4(3):3005--3012.

\bibitem[{Pěnička et~al.(2017)Pěnička, Faigl, Váňa, \protect\BIBand{} Saska}]{Penicka2017}
Pěnička R, Faigl J, Váňa P, Saska M (2017) Dubins orienteering problem. \emph{IEEE Robotics and Automation Letters} 2(2):1210--1217.

\bibitem[{Tawarmalani et~al.(2002)Tawarmalani, Ahmed, \protect\BIBand{} Sahinidis}]{Tawarmalani2002}
Tawarmalani M, Ahmed S, Sahinidis N (2002) Global optimization of 0-1 hyperbolic programs. \emph{Journal of Global Optimization} 24:385--416.

\bibitem[{Train(2003)}]{Trai03}
Train K (2003) \emph{Discrete Choice Methods with Simulation} (Cambridge University Press).

\bibitem[{Tsiogkas \protect\BIBand{} Lane(2018)}]{Tsiogkas2018}
Tsiogkas N, Lane DM (2018) Dcop: Dubins correlated orienteering problem optimizing sensing missions of a nonholonomic vehicle under budget constraints. \emph{IEEE Robotics and Automation Letters} 3(4):2926--2933.

\bibitem[{Vansteenwegen et~al.(2011)Vansteenwegen, Souffriau, \protect\BIBand{} Oudheusden}]{VANSTEENWEGEN2011}
Vansteenwegen P, Souffriau W, Oudheusden DV (2011) The orienteering problem: A survey. \emph{European Journal of Operational Research} 209(1):1--10.

\bibitem[{Vovsha \protect\BIBand{} Bekhor(1998)}]{VovsBekh98}
Vovsha P, Bekhor S (1998) Link-nested logit model of route choice {O}vercoming route overlapping problem. \emph{Transportation Research Record} 1645:133--142.

\bibitem[{Yu et~al.(2022)Yu, Cheng, \protect\BIBand{} Zhu}]{Yu2022}
Yu Q, Cheng C, Zhu N (2022) Robust team orienteering problem with decreasing profits. \emph{INFORMS Journal on Computing} 34(6):3215--3233.

\bibitem[{Zhang et~al.(2012)Zhang, Berman, \protect\BIBand{} Verter}]{Zhang2012}
Zhang Y, Berman O, Verter V (2012) The impact of client choice on preventive healthcare facility network design. \emph{OR Spectrum} 34:349–370.

\end{thebibliography}

\end{document}